%% file: arXiv_EAG.tex
\documentclass[letterpaper]{article} 
\usepackage{aaai2026}  
\usepackage{times}  
\usepackage{helvet}  
\usepackage{courier}  
\usepackage[hyphens]{url}  
\usepackage{graphicx} 
\usepackage{natbib}  
\usepackage{caption} 
\usepackage{algorithm}
\usepackage{algorithmic}
\usepackage{amsmath}
\usepackage{amssymb}
\usepackage{amsthm}

\usepackage{multirow}
\usepackage{graphicx}
\usepackage{subcaption}
\usepackage{colortbl}
\usepackage{xcolor}
\usepackage{booktabs}

\usepackage{overpic}

\usepackage{newfloat}
\usepackage{listings}
\DeclareCaptionStyle{ruled}{labelfont=normalfont,labelsep=colon,strut=off} 
\floatstyle{ruled}
\newfloat{listing}{tb}{lst}{}
\floatname{listing}{Listing}
\title{Energy-based Autoregressive Generation for Neural Population Dynamics}
\author{
    Ningling Ge\textsuperscript{\rm 1,\rm 2,\rm 3}\equalcontrib, 
    Sicheng Dai\textsuperscript{\rm 1,\rm 2,\rm 3,\rm 4}\equalcontrib, 
    Yu Zhu\textsuperscript{\rm 1,\rm 2,\rm 3,\rm 4}\footnote{ Corresponding Author}, 
    Shan Yu\textsuperscript{\rm 1, \rm 3}\textsuperscript{\dag}\\
}
\affiliations{
    \textsuperscript{\rm 1}Institute of Automation, Chinese Academy of Sciences\\
    \textsuperscript{\rm 2}School of Artifcial Intelligence, University of Chinese Academy of Sciences\\
    \textsuperscript{\rm 3}State Key Laboratory of Brain Cognition and Brain-inspired Intelligence Technology\\
    \textsuperscript{\rm 4}Beijing Academy of Artificial Intelligence\\
    geningling2023@ia.ac.cn, daisicheng2023@ia.ac.cn, zhuyu2022@ia.ac.cn, shan.yu@nlpr.ia.ac.cn

}

\usepackage{bibentry}

\begin{document}

\maketitle

\begin{abstract}
Understanding brain function represents a fundamental goal in neuroscience, with critical implications for therapeutic interventions and neural engineering applications. Computational modeling provides a quantitative framework for accelerating this understanding, but faces a fundamental trade-off between computational efficiency and high-fidelity modeling. To address this limitation, we introduce a novel Energy-based Autoregressive Generation (EAG) framework that employs an energy-based transformer learning temporal dynamics in latent space through strictly proper scoring rules, enabling efficient generation with realistic population and single-neuron spiking statistics. Evaluation on synthetic Lorenz datasets and two Neural Latents Benchmark datasets (MC\_Maze and Area2\_bump) demonstrates that EAG achieves state-of-the-art generation quality with substantial computational efficiency improvements, particularly over diffusion-based methods. Beyond optimal performance, conditional generation applications show two capabilities: generalizing to unseen behavioral contexts and improving motor brain-computer interface decoding accuracy using synthetic neural data. These results demonstrate the effectiveness of energy-based modeling for neural population dynamics with applications in neuroscience research and neural engineering. Code is available at https://github.com/NinglingGe/Energy-based-Autoregressive-Generation-for-Neural-Population-Dynamics.
\end{abstract}


\section{Introduction}

Neural population dynamics form a fundamental computational basis of brain function, where coordinated spike patterns across neuron ensembles encode sensory information \cite{romo2003flutter,panzeri2022structures}, motor commands \cite{churchland2012neural,gallego2017neural}, and cognitive states \cite{mante2013context,rigotti2013importance}. Elucidating these computational mechanisms not only provides mechanistic insight into cortical neural coding \cite{gallego2017neural, safaie2023preserved} but also advances therapeutic interventions for neurological disorders including Parkinson's disease \cite{little2013adaptive}, and facilitates brain-computer interfaces (BCIs) for motor restoration in paralysis \cite{hochberg2012reach,willett2023high}.

Computational modeling provides a quantitative framework to analyze neural mechanisms and population dynamics \cite{sussillo2015neural,vyas2020causal}. These approaches typically fall into \textbf{encoding and decoding models} \cite{mathis2024decoding}. Encoding models characterize how external variables are transformed into neural activity patterns by establishing statistical relationships between stimuli and neural responses \cite{walker2019inception,wang2025foundation}, while decoding models take the inverse approach, reconstructing behavioral or stimulus variables from recorded neural activity \cite{gallego2017neural,yoshida2020natural,zhu2025neural}. Decades of work have refined decoders that extract behavior and stimulus features from high-dimensional neural recordings \cite{sussillo2016lfads, ye2023neural, azabou2024multi}. In contrast, encoding models \textbf{remain underexplored}, despite their importance in revealing low-dimensional manifold dynamics \cite{gallego2017neural,pandarinath2018inferring} in motor cortex \cite{gallego2020long,safaie2023preserved,zhu2025neural} and enabling more interpretable, robust decoding for motor BCIs \cite{hochberg2012reach,willett2023high}.

In this framework, neural encoding models can be further divided into predictive and generative models. Predictive models directly map stimuli to neural responses \cite{bashivan2019neural,walker2019inception}, offering efficiency but limited ability to capture \textbf{maintain trial-to-trial variability} \cite{churchland2010stimulus,ecker2014state}. Generative models fall into two main classes: VAE-based methods \cite{zhou2020learning, hurwitz2021targeted, keshtkaran2022large} which sample from latent spaces conditioned on priors but fail to capture complex \textbf{population and single-neuron statistics}; and diffusion-based methods, such as LDNS \cite{kapoor2024latent} and GNOCCHI \cite{mccart2024diffusion}, which model response variability through latent-space distributions but require costly iterative sampling, leading to \textbf{inefficient estimation} of neural statistics.

To efficiently and effectively modeling, we develop a novel Energy-based Autoregressive Generation (EAG) framework. EAG employs an energy-based transformer that learns temporal dynamics in latent space through strictly proper scoring rules \cite{szekely2003statistics}, enabling efficient generation while achieving high fidelity and preserving trial-to-trial variability. The framework supports both unconditional generation for studying neural dynamics and conditional generation for modeling behavior-neural relationships. We evaluate EAG on synthetic Lorenz datasets and two real neural datasets from the Neural Latents Benchmark \cite{pei2021neural}: MC\_Maze and Area2\_bump. The method achieves state-of-the-art (SOTA) generation quality with substantial computational gains, especially delivering a 96.9\% speed-up over diffusion-based approaches. Beyond improvements in generation quality and computational efficiency, conditional generation applications demonstrate two capabilities: (1) generalization to unseen behavioral contexts, revealing generalizable computational mechanisms; and (2) up to a 12.1\% improvement in motor BCI decoding accuracy when trained with EAG-generated data. These results demonstrate the effectiveness of energy-based modeling for neural population dynamics.

In conclusion, our main contributions are as follows:
\begin{itemize}
    \item We develop the \textbf{novel EAG framework} that resolves the trade-off between computational efficiency and high-quality modeling through energy-based learning.
    \item We demonstrate that EAG achieves \textbf{SOTA generation quality} with substantial \textbf{efficiency improvements} over existing methods.
    \item We show that conditional generation enables \textbf{generalization} to unseen behavioral contexts and \textbf{improvement} of motor BCI decoding accuracy, demonstrating practical applications beyond basic neural modeling.
\end{itemize}

\section{Related work}

\textbf{Neural-Behavioral Modeling.} 
Understanding neural computation relies on encoding models that predict neural responses from behavioral variables and decoding models that infer behavioral states from neural signals \cite{mathis2024decoding}. Deep neural networks have advanced encoding models for stimulus-response mappings in visual cortex \cite{yamins2016using,kell2018task,walker2019inception,bashivan2019neural,marks2021stimulus,vargas2024task,wang2025foundation} and decoding models for extracting behavioral information from motor \cite{gallego2017neural} and visual cortices \cite{yoshida2020natural,stringer2021high}. However, these approaches primarily focus on input-output mappings cannot capture the stochastic variability inherent in neural population dynamics.

\textbf{Neural Spike Generation.} 
Neural spike generation addresses data scarcity in brain-computer interface applications through synthetic data augmentation for decoder training and stability \cite{wen2023rapid,ma2023using}. Variational autoencoders and generative adversarial networks have been applied to neural population modeling and spike train synthesis \cite{pandarinath2018inferring,wen2023rapid,ma2023using}. Latent diffusion for neural spike generation \cite{kapoor2024latent,mccart2024diffusion} encodes spike trains into continuous latent spaces and applies diffusion models to generate behaviorally-conditioned neural activity, but requires computationally expensive iterative denoising steps \cite{ho2020denoising}. In contrast, this work applies energy-based modeling to autoregressive neural spike generation, enabling efficient sampling while maintaining high-quality spike pattern generation.

\textbf{Energy-based Models.}
Energy-based models define probability distributions through energy functions, where probability is inversely related to energy \cite{hinton2002training,lecun2006tutorial}. These models enable direct sampling from learned distributions, providing a principled approach to generative modeling \cite{song2019generative}. EBMs have been successfully applied across diverse domains including image generation \cite{song2019generative}, natural language processing \cite{bakhtin2021residual}, molecular design \cite{satorras2021n}, and protein structure design \cite{watson2023novo}. In contrast, this work introduces EBMs to neural computational modeling for the first time.

\section{Preliminaries}
\subsection{Strictly Proper Scoring Rules}
Scoring rules evaluate probabilistic forecasts by comparing predicted distributions against observed outcomes. Given sample space $\mathcal{X}$ and probability measures $\mathcal{P}$ on $\mathcal{X}$, a scoring rule $S$ maps predicted distributions $p$ and observed samples $x$ to extended real values:
\begin{equation}
 S(p,x):\mathcal{P} \times \mathcal{X} \mapsto \overline{\mathbb{R}}.
\end{equation}
The expected score under the true distribution $q$ quantifies prediction quality:
\begin{equation}
\label{eq:exp_score}
S(p,q)=\mathbb{E}_{x\sim q}[S(p,x)].
\end{equation}
A scoring rule is proper if truthful reporting maximizes expected scores:
\begin{equation}
S(p,q) \leq S(q,q), \ \ \ \  \forall p,q \in \mathcal{P}.
\end{equation}
Strict propriety occurs when equality holds exclusively for $p=q$, ensuring unique optimal predictions.

Classical scoring rules include the Brier score \citep{brier1950statistical}, logarithmic score \citep{good1952rational} and spherical score \citep{roby1965belief}. For continuous distributions, the energy score \citep{szekely2003statistics} provides a strictly proper scoring rule that will be utilized in this work.

\subsection{Scoring Rules for Generative Modeling}
Strictly proper scoring rules provide objectives for training generative models through negative score loss functions:
\begin{equation}
\label{eq:l}
\mathcal{L}_S(p,x)=-S(p,x).
\end{equation}
The logarithmic score yields cross-entropy loss when maximized. Strict propriety ensures unique optimization targets.

For sequential data, autoregressive generation decomposes the loss across time steps:
\begin{equation}
\label{eq:l_auto}
\mathcal{L}_S(p,x)=-\sum_{t=1}^{T}S(p(\cdot|x_{<t}),x_t).
\end{equation}
Expected loss minimization requires each conditional distribution $p(\cdot|x_{<t})$ to match the true conditional $q(\cdot|x_{<t})$.

Neural spike data lacks explicit likelihood forms, resulting direct score calculation intractable. This necessitates tractable score estimators that preserve strict propriety for neural spike generation.


\begin{figure*}[tb]
\centering
\includegraphics[scale=1.5]{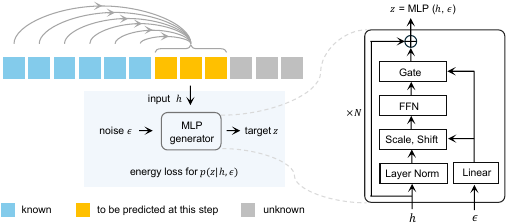}
\caption{Energy-based Autoregressive Generation (EAG) framework. Known latent positions (blue) provide context for predicting masked positions (gray). The MLP generator incorporates noise $\boldsymbol{\epsilon}$ via adaptive layer normalization to enable stochastic generation, trained with energy loss for distributional prediction.}
\label{fig:method}
\end{figure*}

\section{Methods}

\subsection{Latent Generation Framework}

EAG adopts a two-stage paradigm that first learns compact neural representations and then performs efficient generation in the latent space. While Stage 1 follows established autoencoder approaches for fair comparison, EAG's core contribution lies in Stage 2, where we introduce a novel energy-based generation mechanism that fundamentally differs from existing diffusion-based methods.

\textbf{Stage 1: Neural Representation Learning.} Following LDNS \cite{kapoor2024latent}, we employ their autoencoder architecture to obtain latent representations $\mathbf{z} \in \mathbb{R}^{d \times T}$ from neural spiking data $\mathbf{s} \in \mathbb{N}_0^{n \times T}$ and optional behavioral covariates $\mathbf{y}$, where $d \ll n$. This stage maps high-dimensional spike trains to a low-dimensional latent space under a Poisson observation model with temporal smoothness constraints. We use identical network architecture and training configuration as \cite{kapoor2024latent}.

\textbf{Stage 2: Energy-based Latent Generation.} Given the learned latent representations $\mathbf{z}$ from Stage 1, EAG employs an energy-based autoregressive framework for latent generation. The approach predicts missing latent representations through masked autoregressive modeling guided by the energy score, a strictly proper scoring rule that does not require explicit likelihood computation. This formulation enables single-pass generation while preserving stochastic properties necessary for modeling trial-to-trial variability. The detailed methodology is presented in the following section.

\subsection{Energy-based Autoregressive Generation}

Building upon the latent representations $\mathbf{z}$ learned in Stage 1, EAG employs energy-based modeling for neural population dynamics generation through autoregressive prediction, as shown in Figure \ref{fig:method}. The framework addresses a fundamental challenge in neural generative modeling: predicting distributions over continuous latent spaces without explicit likelihood computation while preserving the stochastic nature of neural variability.

\subsubsection{Energy Loss}

The energy score provides a strictly proper scoring rule for continuous latent variables in $\mathbb{R}^d$. For model distribution $p_{\theta}$ generating latent samples $\mathbf{z}$ and data distribution with ground truth latents $\mathbf{z}_{\text{data}}$, the energy score with parameter $\alpha \in (0,2)$ is defined as:

\begin{equation}
S(p_{\theta},\mathbf{z}_{\text{data}}) = \mathbb{E}[\|\mathbf{z}_1-\mathbf{z}_2\|^{\alpha}] - 2 \mathbb{E}[\|\mathbf{z}-\mathbf{z}_{\text{data}}\|^{\alpha}]
\end{equation}

where $\mathbf{z}_1, \mathbf{z}_2, \mathbf{z}$ are independent samples from $p_{\theta}$. The energy score achieves strict propriety for $\alpha \in (0,2)$, ensuring that optimal predictions correspond to the true latent distribution.

The energy loss can be unbiasedly estimated using two independent latent samples $\mathbf{z}_1, \mathbf{z}_2$ from the model distribution:

\begin{equation}
\label{eq:energy_loss}
\mathcal{L}_{\text{energy}}(p_{\theta},\mathbf{z}_{\text{data}})=\|\mathbf{z}_1-\mathbf{z}_{\text{data}}\|^{\alpha} + \|\mathbf{z}_2-\mathbf{z}_{\text{data}}\|^{\alpha} - \|\mathbf{z}_1-\mathbf{z}_2\|^{\alpha}
\end{equation}

This formulation performs distributional prediction by balancing two objectives: the first two terms minimize prediction error while the third term maintains sample diversity. Essentially, the energy loss trains the model to generate samples that are both accurate and appropriately variable, capturing the distributional properties necessary for modeling neural trial-to-trial variability. Unlike diffusion models that require iterative sampling steps, this approach enables direct distributional prediction in a single forward pass while maintaining generative capabilities.

\subsubsection{Energy Transformer Architecture}

The energy transformer architecture enables stochastic latent generation through noise-conditioned output layers while maintaining standard transformer processing for temporal dynamics modeling.

\textbf{Input Processing.} Latent representations $\mathbf{z} \in \mathbb{R}^{d}$ are mapped to model dimension $d_{\text{model}}$ through linear projection layers. Positional encodings are applied to maintain temporal ordering information across the latent sequence.

\textbf{Stochastic Output Generation.} As illustrated in Figure \ref{fig:method}, the output layer incorporates random noise $\boldsymbol{\epsilon}$ through a multi-layer perceptron generator to enable stochastic latent generation. The noise vector of dimension $d_{\text{noise}}$ is sampled from uniform distribution $[-0.5, 0.5]$ and embedded to dimension $d_{\text{mlp}}$ through learned linear transformations.

The MLP generator employs residual blocks with adaptive layer normalization to inject noise into latent predictions, as shown in the architectural detail of Figure \ref{fig:method}. For the $i$-th residual block with input $\mathbf{h}^i$:

\begin{equation}
\begin{aligned}
&\mathbf{h}^i_{\boldsymbol{\epsilon}}=(1+\text{scale}(\boldsymbol{\epsilon}))\cdot \text{LN}(\mathbf{h}^i) + \text{shift}(\boldsymbol{\epsilon})\\
&\mathbf{h}^{i+1} = \mathbf{h}^i + \text{gate}(\boldsymbol{\epsilon}) \cdot \text{FFN}(\mathbf{h}^i_{\boldsymbol{\epsilon}})
\end{aligned}
\end{equation}
where $\text{shift}(\cdot)$, $\text{scale}(\cdot)$, and $\text{gate}(\cdot)$ are learned linear transformations that interpret noise input as adaptive parameters for controlling the scale and bias of feature transformations. The gating mechanism allows the model to selectively incorporate stochastic variations based on the current context.

\textbf{Masked Autoregressive Generation.} As depicted in Figure \ref{fig:method}, the framework employs masked autoregressive modeling where known latent positions (blue) provide temporal context for predicting masked positions (gray). During training, random masking ratios sampled uniformly from $[0.7, 1.0]$ are applied to latent sequences, ensuring the model learns to handle various degrees of missing information. The masking strategy randomly selects time points to predict while preserving causal relationships in the temporal sequence.

During inference, latent time points are generated progressively with masking ratio decreasing from 1.0 to 0 following a cosine schedule. This approach enables bidirectional attention during training for improved context utilization while maintaining autoregressive properties during generation. The progressive unmasking allows the model to iteratively refine predictions based on previously generated latent states.

\subsubsection{Conditional Generation} 
For behavior-conditioned neural generation, EAG incorporates behavioral covariates $\mathbf{y}$ into the input sequence as conditioning tokens. Behavioral variables are embedded through a linear projection layer to match the latent dimension, then concatenated with the latent representations.

During training, behavioral conditions are randomly replaced with learnable null tokens for 10\% of trials to enable classifier-free guidance. This dropout strategy allows the model to learn both conditional and unconditional generation within a unified framework. 

At inference, the model generates latent representations for both the given behavioral condition $\mathbf{h}_c$ and null condition $\mathbf{h}_u$, with the final representation computed using classifier-free guidance:
\begin{equation}
\mathbf{h} = \gamma \cdot \mathbf{h}_c + (1-\gamma) \cdot \mathbf{h}_u
\end{equation}
where $\gamma$ controls the strength of behavioral conditioning. This enables flexible control over the degree of behavioral constraint during generation.

\section{Results}
Our experiments proceed in four stages. First, we conduct unconditional generation on three diverse distinct datasets, and benchmark its performance against VAE-based and diffusion-based generative models on two real neural datasets. Second, we demonstrate that EAG achieves 30× higher sampling efficiency than diffusion models, enabling faster generation without sacrificing quality. Third, EAG generalizes well to unseen labels on conditional generation, capturing neural-behavioral relationships while preserving trial variability. Finally, we show that EAG enhances downstream BCI decoding performance.

\begin{figure}[tb]
\centering
\includegraphics[scale=0.28]{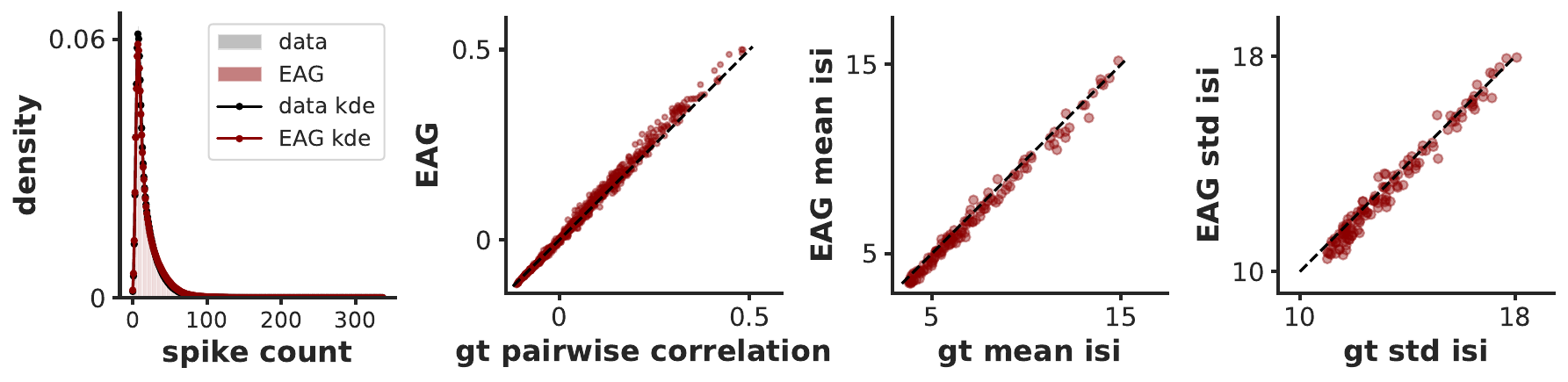}
\caption{\textbf{Unconditional generation on Lorenz Dataset}. The generated data closely matches the ground truth across four metrics: spike count distribution, pairwise correlation, mean-isi, and std-isi.}
\label{fig:lorenz_metrics}
\end{figure}

\textbf{Datasets}. We evaluate our method on one synthetic dataset and two real neural datasets. The synthetic dataset is generated by simulating 128-dimensional neural spiking activity using a Lorenz attractor. The two real datasets, MC\_Maze and Area2\_Bump, are from the Neural Latents Benchmark \cite{pei2021neural}, which record neural activity from different cortical areas of macaque monkeys performing various tasks. Detailed descriptions are provided in the supplementary.

\textbf{Baselines}. To rigorously benchmark the generative performance of EAG, we compared it against a suite of established baseline models drawn from both VAE-based and diffusion-based models. The VAE-based baselines include Targeted Neural Dynamical Modeling (TNDM) \cite{hurwitz2021targeted}, Poisson-identifiable VAE (pi-VAE) \cite{zhou2020learning}, and AutoLFADS \cite{sussillo2016lfads,pandarinath2018inferring, sedler2023lfads,keshtkaran2022large}. Additionally, we include Latent Diffusion for Neural Spiking (LDNS) \cite{kapoor2024latent} as a diffusion-based baseline, given its recent advances over AutoLFADS. Detailed settings are provided in the supplementary.

\textbf{Metrics}. For evaluation, we adopted the comprehensive set of metrics previously proposed in the LDNS study \cite{kapoor2024latent}. The four metrics include population spike count distribution, pairwise spike-count correlations, mean inter-spike interval (ISI), and ISI standard deviation, capturing both population-level and single-neuron spiking statistics. 

\begin{table*}[htb]
    \centering
    \begin{tabular}{c|c|cccc}
        \toprule
         Dataset  &  Method & $D_{KL}$ psch & RMSE pairwise corr & RMSE mean isi & RMSE std isi \\
         \midrule
         \multirow{5}{*}{MC\_Maze} 
            & TNDM & \underline{0.0028 $\pm$ 6.0e-5} & 0.0027 $\pm$ 1.2e-5 & 0.057 $\pm$ 0.004 & 0.029 $\pm$ 0.001 \\
            & pi-VAE & 0.0063 $\pm$ 2.0e-4 & 0.0031 $\pm$ 1.1e-5 & 0.064 $\pm$ 0.002 & 0.034 $\pm$ 0.001 \\
            & AutoLFADS & 0.0040 $\pm$ 2.2e-4 & 0.0026 $\pm$ 1.3e-5 & 0.039 $\pm$ 0.003 & 0.029 $\pm$ 0.001 \\
            & LDNS & 0.0039 $\pm$ 3.0e-4 & \underline{0.0025 $\pm$ 1.1e-4} & \underline{0.037 $\pm$ 0.001} & \underline{0.023 $\pm$ 0.0001} \\
            & \textbf{EAG} & \textbf{0.0014} $\pm$ \textbf{2.0e-4} & \textbf{0.0024} $\pm$ \textbf{1.0e-5} & \textbf{0.024} $\pm$ \textbf{0.001} & \textbf{0.018} $\pm$ \textbf{0.0024} \\
        
         \midrule
        \multirow{5}{*}{Area2\_Bump} 
            & TNDM & 0.0027 $\pm$ 2.9e-4 & 0.0077 $\pm$ 1.0e-4 & 0.049 $\pm$ 0.009 & 0.039 $\pm$ 0.003 \\
            & pi-VAE    & 0.0067 $\pm$ 4.2e-4 & 0.0088 $\pm$ 7.9e-5 & 0.050 $\pm$ 0.007 & \underline{0.029 $\pm$ 0.004} \\
            & AutoLFADS & 0.0032 $\pm$ 3.2e-4 & 0.0081 $\pm$ 1.2e-5 & \underline{0.048 $\pm$ 0.003} & 0.031 $\pm$ 0.006\\
            & LDNS      & \underline{0.0020 $\pm$ 1.2e-4} & \underline{0.0076 $\pm$ 1.4e-5} & 0.050 $\pm$ 0.002 & 0.034 $\pm$ 0.002 \\
            & \textbf{EAG} & \textbf{0.0018} $\pm$ \textbf{1.6e-4} & \textbf{0.0075 $\pm$ 9.1e-5} & \textbf{0.035} $\pm$ \textbf{0.004} & \textbf{0.025} $\pm$ \textbf{0.003} \\
         \bottomrule
    \end{tabular}
    \caption{\textbf{Model metrics comparison}. $D_{KL}$ for the population spike count histogram and RMSE comparisons. Results are reported as mean $\pm$ standard deviation over 5 folds. EAG outperforms all baselines (Wilcoxon, $p < 0.001$), except for pairwise correlation vs. LDNS (n.s., $p = 0.09$). Bold indicates the best-performing model; underlined values indicate the second best.}
    \label{tab:metrics}
\end{table*}

\begin{figure}[t]
\centering
\includegraphics[scale=0.4]{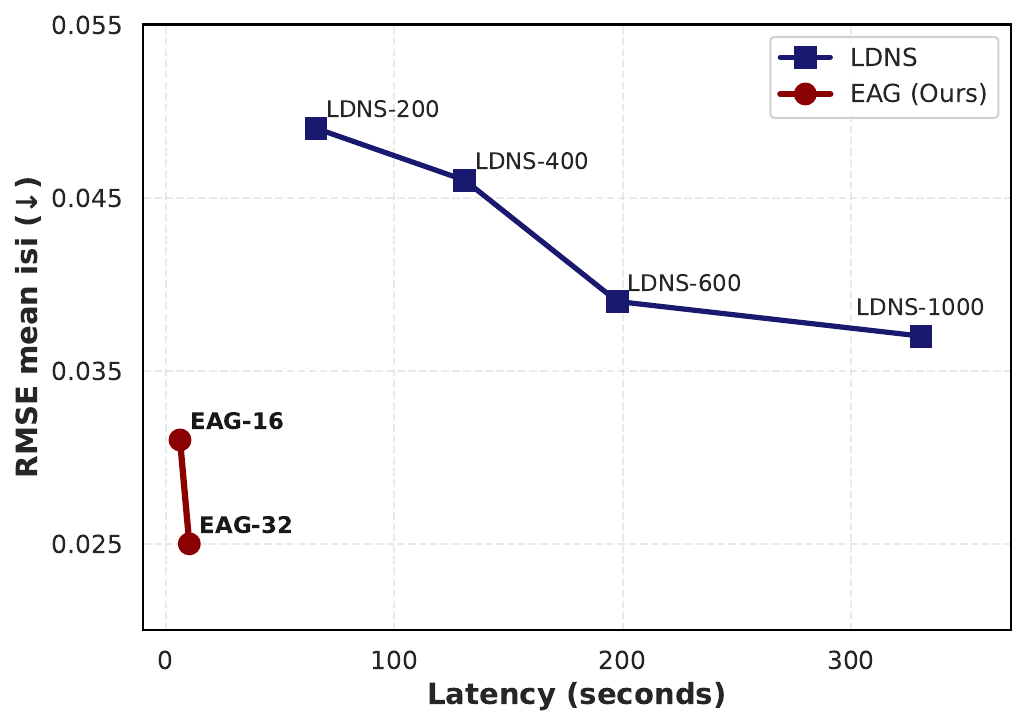}
\caption{\textbf{The latency/quality trade-off for EAG and LDNS.} We vary number of diffusion steps (200, 400, 600, 1000) of LDNS and number of autoregressive steps (16, 32) of EAG. EAG-32 achieves a 96.9\% reduction in latency, and a 32.4\% improvement on RMSE mean ISI compared to LDNS-1000.}
\label{fig:eff_meanisi}
\end{figure}

\subsection{EAG generates high-quality neural spike data}

To validate effectiveness of EAG framework, we first evaluate it under an unconditional generation setting on three datasets characterized by diverse distributions and high variability. We demonstrate that EAG achieves state-of-the-art performance against both VAE-based and diffusion-based methods.

\textbf{Lorenz Dataset}. We first apply EAG to a synthetic dataset generated from a 3D Lorenz system, where 128-dimensional neural observations over 256 time steps are derived from projections of the underlying attractor dynamics. In the first stage, we train an autoencoder to effectively reconstruct the firing rates. Then we train EAG within the latent space in the second stage. Supp. Figure S1 and Figure S2 visualize Lorenz rates generated by EAG and the corresponding spike trains obtained via Poisson sampling, which are visually indistinguishable from the real data. We quantitatively evaluate EAG’s performance, as visualized in Figure \ref{fig:lorenz_metrics}. EAG captures both population-level features (e.g., spike count distribution, pairwise correlation) and single-neuron statistics (e.g., ISI mean and variance). Additionally, Supp. Figure S3 demonstrates that both AE-reconstructed and EAG-generated rates accurately capture the ground-truth pairwise correlation structure of the synthetic Lorenz rates.

\textbf{MC\_Maze Dataset}. We then benchmark EAG on MC\_Maze \cite{churchland2022mc_maze, pei2021neural}, a widely used real neural dataset often referred to as the "neural MNIST," recorded from premotor and primary motor cortex as a monkey performs a delayed center-out reaching task with spatial barriers. EAG is shown to generate highly realistic sparse spike trains (Supp. Figure S4). Quantitative comparison against VAE-based and diffusion-based baselines demonstrates that EAG consistently achieves the best performance across all four evaluation metrics as Table \ref{tab:metrics} (visualization in Supp. Figure S5). Notably, even after augmenting baselines with spike-history inputs (a trick known to benefit LDNS), EAG still achieves the best performance (Supp. Table S1). In addition, a plain autoregressive Transformer trained with MSE loss performs significantly worse than EAG (Supp. Table S1), highlighting the essential contribution of the energy loss.

\textbf{Area2\_Bump Dataset}. To further test EAG’s robustness on limited data and non-autonomous neural activity, we evaluate it on the Area2\_Bump dataset, a small-scale neural dataset containing about 300 trials recorded from the somatosensory cortex during a bump-perturbed reaching task. Despite the data scarcity, EAG continues to generate spike trains that are both visually and statistically aligned with the real data (Supp. Figure S6, Supp. Figure S7), outperforming all baselines as Table \ref{tab:metrics} shows. 
Additionally, Supp. Table S2 shows that EAG maintains the best performance on all evaluation metrics compared to spike-history-augmented versions.

Collectively, these results highlight EAG’s strong capacity to model both population-level and single-neuron-level patterns and generate biologically realistic spike data, with stable performance across varying data scales and cortical regions.

\begin{figure*}[htb]
    \centering
    \begin{subfigure}[t]{0.235\textwidth}
        \centering
        \includegraphics[width=\linewidth]{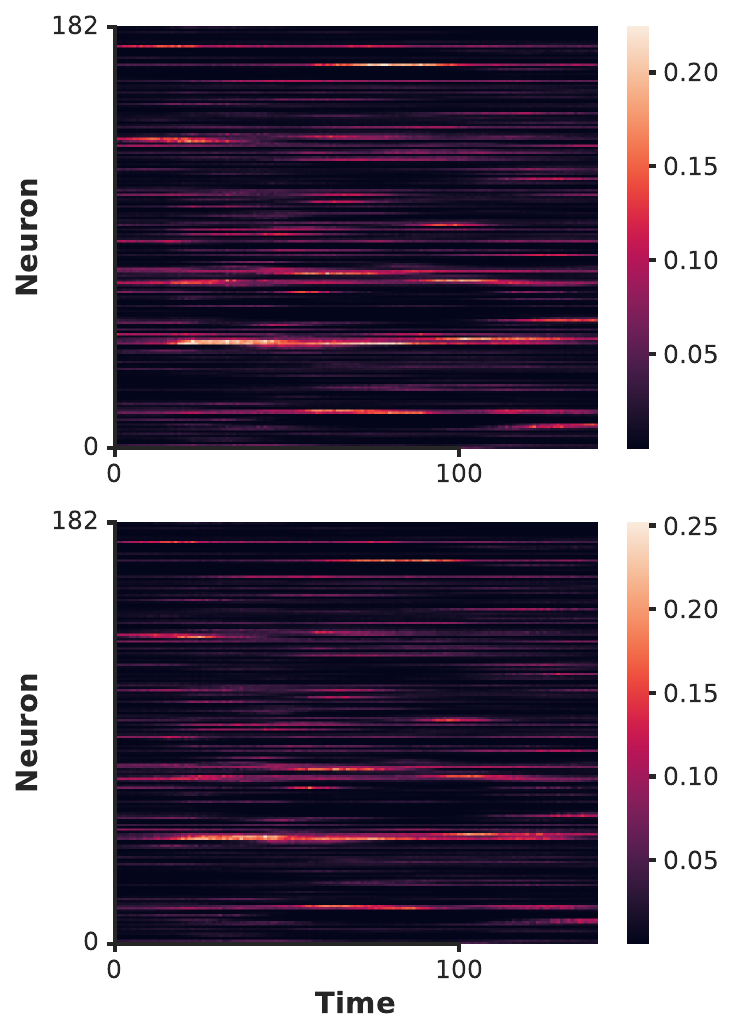}
        \caption{Firing rates}
        \label{fig:angcond_fr}
    \end{subfigure}
    \hspace{0.015\textwidth}
    \begin{subfigure}[t]{0.24\textwidth}
        \centering
        \includegraphics[width=\linewidth]{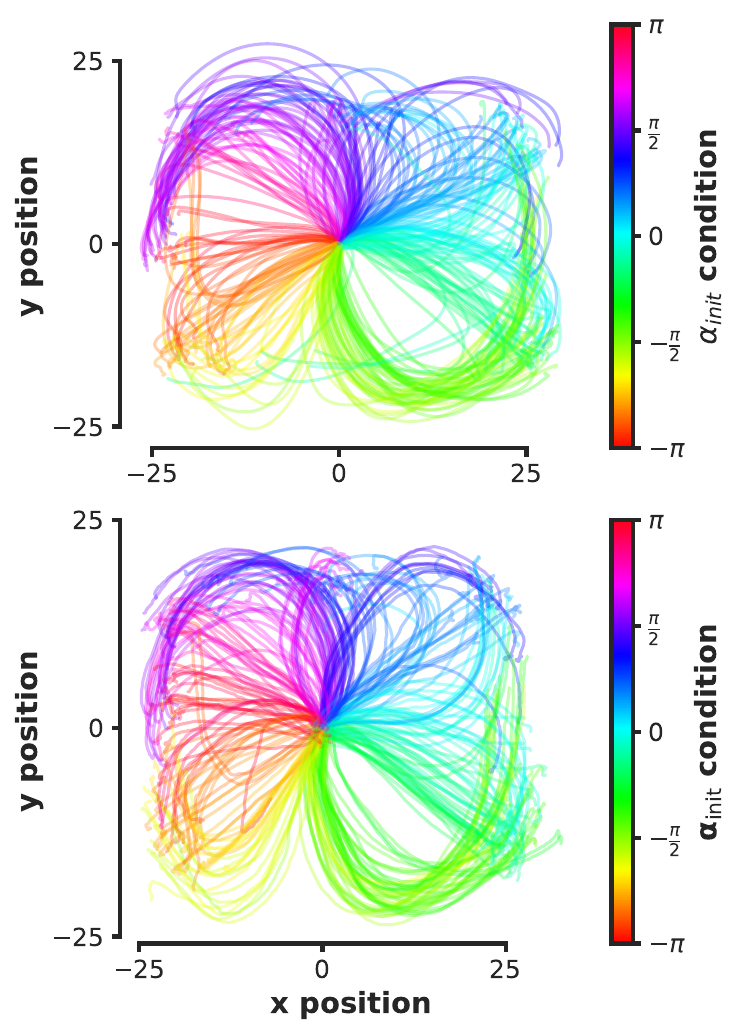}
        \caption{Decoded trajectory}
        \label{fig:angcond_traj}
    \end{subfigure}
    \hspace{0.015\textwidth}
    \begin{subfigure}[t]{0.23\textwidth}
        \centering
        \includegraphics[width=\linewidth]{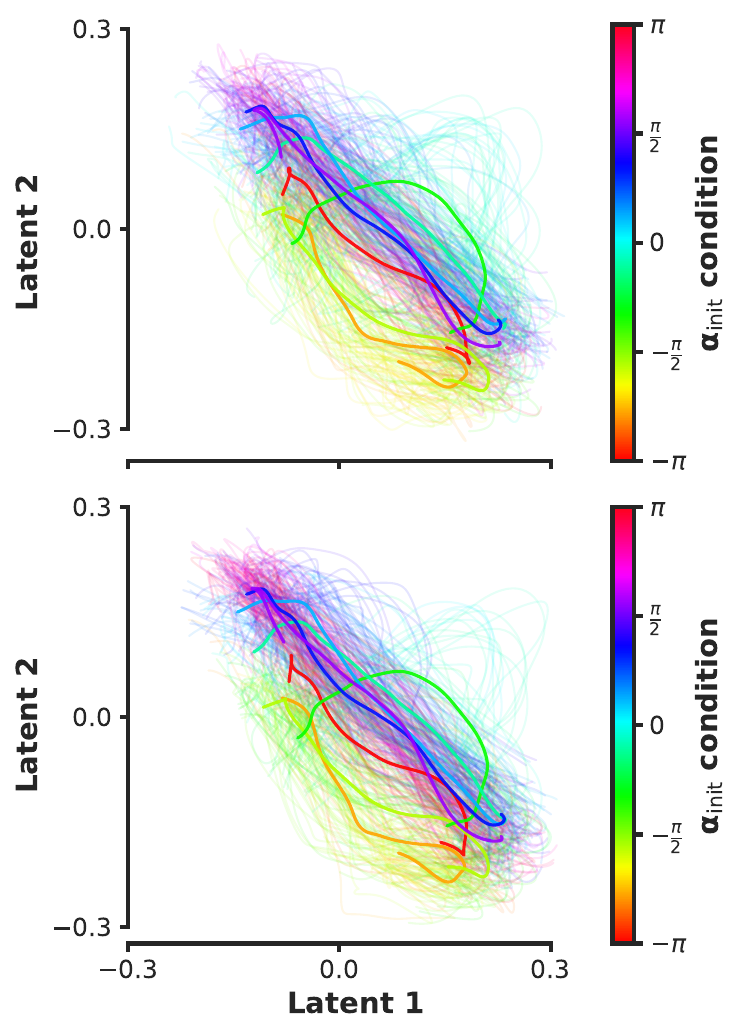}
        \caption{Single-trial neural latents}
        \label{fig:angcond_latent}
    \end{subfigure}
    \hspace{0.015\textwidth}
    \begin{subfigure}[t]{0.185\textwidth}
        \centering
        \includegraphics[width=\linewidth]{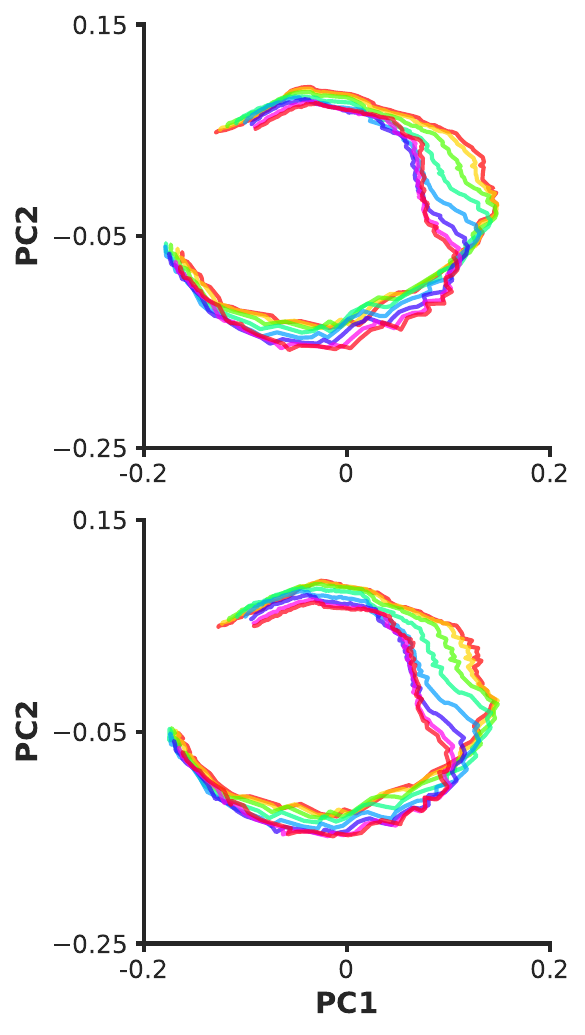}
        \caption{Averaged neural PCs}
        \label{fig:angcond_pc}
    \end{subfigure}
    
    \caption{\textbf{Generalization to unseen angle labels.} (a) real firing rates and sampled firing rates. (b) Decoded trajectory from real rates and sampled rates conditioned on unseen angle labels. (c) Single-trial neural trajectories in latent space extracted from real and sampled activity. (d) The first 2 principal components averaged over eight reach directions of real and sampled firing rates. For all panels, top: real data; bottom: sampled data.}
    \label{fig:angcond}
\end{figure*}

\subsection{EAG samples with high efficiency}
To assess the efficiency of EAG in addition to its generation quality, we conducted a systematic comparison with diffusion-based models. Unlike LDNS, which relies on iterative denoising steps during inference, EAG generates in a single forward pass, resulting in orders-of-magnitude improvements in sampling speed. We trained four LDNS models with diffusion steps ranging from 200 to 1000, and two EAG models with autoregressive steps 16 and 32, and evaluated both generation quality and inference latency. Here, latency refers to the time taken to generate the same number of trials as in the training dataset. ($\sim$2000 trials).

Taking single-neuron generation quality as a representative case, Figure \ref{fig:eff_meanisi} illustrates the relationship between RMSE mean-isi and generation latency across varying steps of LDNS and EAG. In terms of sample latency, EAG-32 generates 2008 trials in just 10.29s while LDNS-1000 requires 330.64s, achieving a 96.9\% speed‐up. Against the minimal-step LDNS-200, EAG-32 still achieves an 84.4\% reduction in latency. In terms of generation quality, even minimal-step EAG-16 outperforms LDNS-1000. Specifically, EAG-32 delivers a 49.0\% gain over LDNS-200 and a 32.4\% gain over LDNS-1000. Similar trends are also observed across other quality metrics, as detailed in Supp. Figure S8 and Supp. Table S3. Moreover, scaling analysis further shows that EAG’s runtime and memory usage remain largely unaffected by increases in neuron count or time length (details in Supp.Figure S4). These results demonstrate that energy-based autoregressive generation achieves higher efficiency than diffusion-based methods while maintaining better quality and stable to scaled datasets.

\subsection{EAG generalizes to unseen contexts}

EAG’s modeling capability generalizes effectively to unseen behavioral contexts. To systematically evaluate this property, we condition EAG on behavior variables not observed during training and assess whether the model can generate realistic neural activity consistent with the given context. 

\begin{figure}[h]
    \centering
    \begin{subfigure}{\linewidth}
        \centering
        \begin{overpic}[scale=0.27]{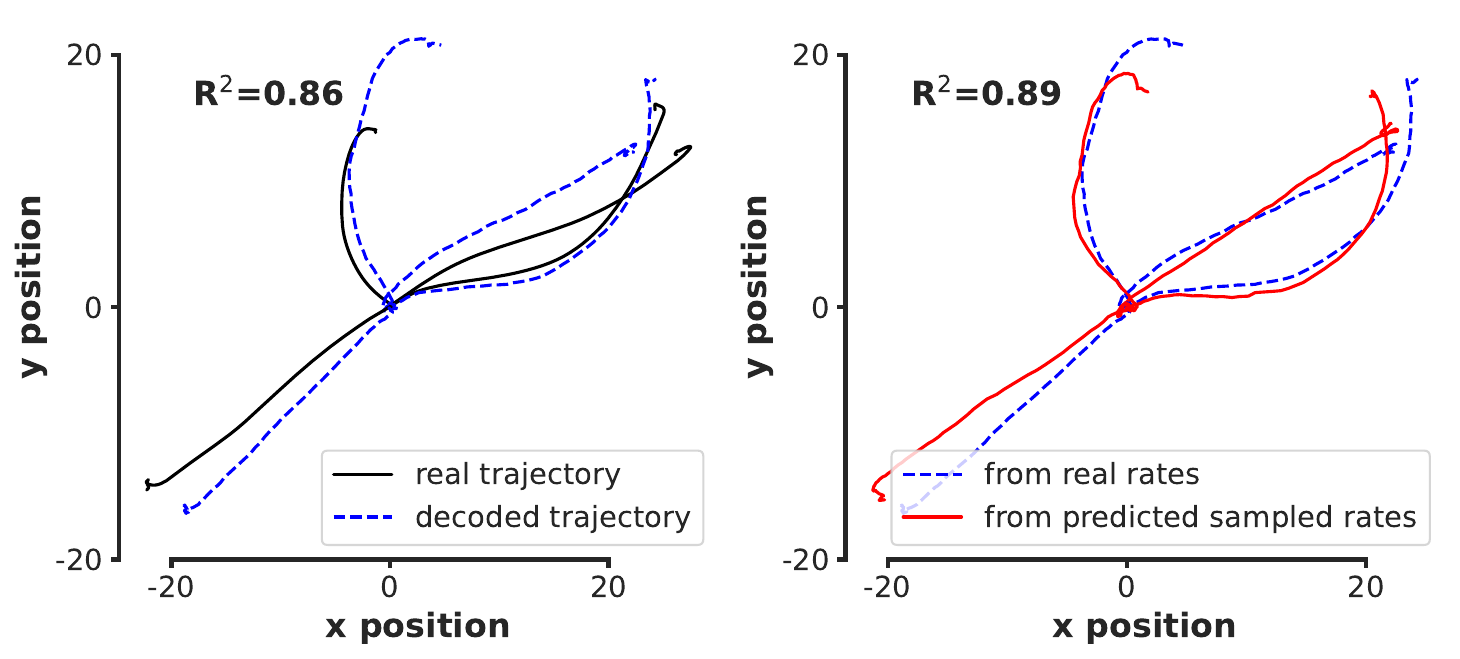}
            \put(-15,47){\small (a)}
        \end{overpic}
    \end{subfigure}

    \begin{subfigure}{\linewidth}
        \centering
        \begin{overpic}[scale=0.27]{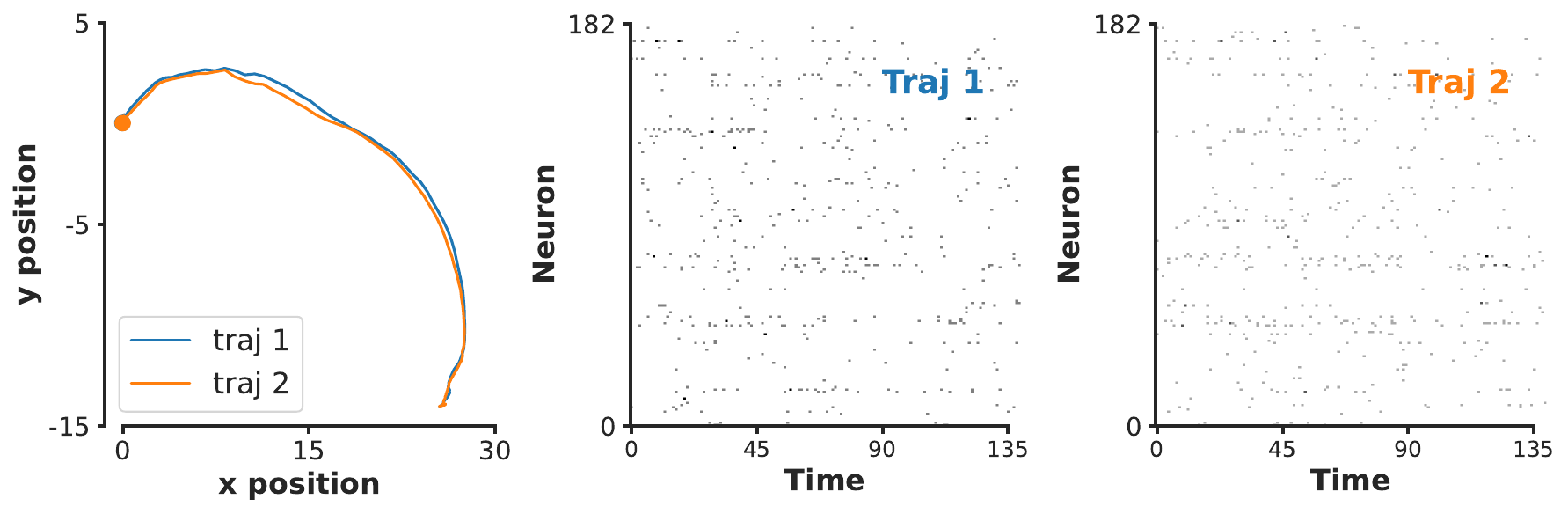}
            \put(-3,33){\small (b)}
        \end{overpic}
    \end{subfigure}

    \caption{\textbf{Generalization to unseen velocity labels.} (a) Real hand trajectory and decoded trajectory (left panel), decoded trajectory from real rates and sampled rates (right panel). (b) Sampled spike trains conditioned on two nearly identical trajectories show trial-to-trial variability.}
    \label{fig:velcond}
\end{figure}

We first condition EAG on the \textbf{monkey’s initial reach direction} $\alpha_{\text{init}}$. Figure \ref{fig:angcond_fr} shows real rates (top panel) and generated rates (bottom panel) conditioned at a novel angles ($\sim60^\circ$), which align closely (additional samples in Supp. Figure S10). The generated activity is not a direct copy but shows natural trial-to-trial variability. EAG's ability to model variability is further examined in velocity-conditioned generation. Next, using the closed-loop evaluation pipeline described in Supp. Figure S9, we decode trajectories via a pretrained ridge regression model from EAG-sampled rates conditioned on unseen labels. As shown in Figure \ref{fig:angcond_traj}, the decoded (bottom) and true trajectories (top) closely match. To verify that EAG effectively captures the underlying mechanisms of neural-activity relations, we visualize real and sampled single-trial latents from the EAG-extracted latent space (Figure \ref{fig:angcond_latent}). Despite unseen labels, the latent distributions clearly align with angle labels, matching real-data patterns. We further apply PCA to the high-dimensional firing rates and average PCs for eight reach directions in a 5D space, which shows highly similar real and generated trajectories (Figure \ref{fig:angcond_pc}). These results demonstrate EAG’s capacity to model latent neural–behavioral structure, enabling generalization to unseen conditions.

Then we consider a more fine-grained behavioral label by conditioning EAG on the hand velocity $(v_x, v_y)$ at each time point. EAG produces firing rates that closely resemble the real ones when conditioned on an entirely unseen velocity trajectory, as shown in Supp. Figure S11. Figure \ref{fig:velcond}a shows the ridge model bias (left panel) when decoding from real rates, while the right panel shows trajectories decoded from EAG-sampled rates, achieving a higher similarity with the ground truth ($R^2=0.89$) compared to LDNS ($R^2=0.65$). Notably, even for two trajectories that are virtually identical, EAG-sampled spike trains retain strong trial-to-trial variability (Figure \ref{fig:velcond}b). This ability to generate realistic neural activity from hypothetical movements while preserving variability are vital for downstream applications such as BCI decoding \cite{wen2023rapid, degenhart2020stabilization}.

\subsection{EAG enhances BCI performance}

To assess the practical impact of EAG-generated data, we investigate its effectiveness in improving BCI decoding performance through data augmentation. Specifically, using two Neural Latent Benchmark datasets, MC\_Maze and Area2\_Bump, we evaluate changes in decoding accuracy for multiple baseline models when trained with and without onefold EAG-based augmentation. As shown in Table \ref{tab:mcmaze_aug}, EAG consistently improves decoding on MC\_Maze, with larger gains in more complex models. In particular, the transformer-based Neural Data Transformer (NDT) exhibits the most pronounced gain, reaching up to 12.1\%. To ensure that these improvements are not artifacts of suboptimal training, we further apply ray-based hyperparameter optimization for both LFADS and NDT (autoLFADS and NDT-ray), and compare performance with and without augmentation. The improvements are reduced but persist, validating effectiveness and the robustness of EAG’s augmentation benefits. 

\begin{table}[t]
    \centering
    \setlength{\tabcolsep}{1mm}
    \scalebox{0.74}{
        \begin{tabular}{c|c c c}
        \toprule
         Method     &  Co-smoothing bps($\uparrow$)  &  Behavior decoding($\uparrow$)  &  PSTH $R^2$($\uparrow$) \\
         \midrule
         GRU        &  0.267/$\mathbf{0.270}$(+1.1\%) & 0.868/$\mathbf{0.880}$(+1.3\%) & 0.507/$\mathbf{0.533}$(+5.1\%)  \\
         SLDS        &  0.218/$\mathbf{0.233}$(+6.9\%) & 0.794/$\mathbf{0.801}$(+0.8\%) & 0.478/$\mathbf{0.492}$(+3.0\%)  \\
         LFADS       &  0.324/$\mathbf{0.347}$(+7.1\%) & 0.901/$\mathbf{0.906}$(+0.5\%) & 0.579/$\mathbf{0.594}$(+2.6\%)  \\
         NDT        &  0.272/$\mathbf{0.305}$(+12.1\%) & 0.776/$\mathbf{0.838}$(+8.0\%) & 0.546/$\mathbf{0.581}$(+6.4\%)   \\
        \midrule
         AutoLFADS   &  0.347/$\mathbf{0.350}$(+0.9\%) & 0.906/$\mathbf{0.911}$(+0.6\%) & 0.594 /$\mathbf{0.607}$(+2.2\%)  \\
         NDT(ray)   &  0.335/$\mathbf{0.363}$(+8.4\%)  & 0.875/$\mathbf{0.904}$(+3.3\%) & 0.589/$\mathbf{0.596}$(+1.2\%)  \\
         \bottomrule
        \end{tabular}
    }
    \caption{\textbf{Metrics before and after EAG augmentation on MC\_Maze.} Metrics follow the NLB evaluation and are formatted as before/after(improvement ratio). Bold indicates improvement after EAG augmentation.}
    \label{tab:mcmaze_aug}
\end{table}

Similar results are observed on the smaller Area2\_Bump dataset (Supp. Table S5). Given its limited size relative to MC\_Maze, we hypothesize that larger scale augmentation brings more decoding gains. We test the effect of scaling EAG augmentation to 1$\times$, 2$\times$, and 4$\times$, using NDT as the decoder. As shown in Table \ref{tab:area2bump_multi_aug}, accuracy rises as data scale increased, with the largest jump yielding up to a 54.7\% improvement at 2$\times$ augmentation in NDT. This confirms that, for small datasets, data scarcity represents a more pressing bottleneck in decoding tasks, and increased augmentation yields greater gains.

\begin{table}[t]
    \centering
    \setlength{\tabcolsep}{1mm}
    \scalebox{0.69}{
        \begin{tabular}{c|c c c}
        \toprule
         Method     &  Co-smoothing bps($\uparrow$)  &  Behavior decoding($\uparrow$)  &  PSTH $R^2$($\uparrow$) \\
         \midrule
         NDT-2size        &  0.106/$\mathbf{0.164}$(+54.7\%) &  0.512/$\mathbf{0.621}$(+21.3\%) & 0.321/$\mathbf{0.422}$(+31.5\%) \\
         NDT-4size        &  0.106/$\mathbf{0.161}$(+51.9\%) &  0.512/$\mathbf{0.631}$(+23.2\%) & 0.321/$\mathbf{0.425}$(+32.4\%) \\
         NDT(ray)-2size   &  0.208/$\mathbf{0.227}$(+9.7\%)  &  0.794/$\mathbf{0.854}$(+7.6\%)  & 0.414/$\mathbf{0.499}$(+20.5\%) \\
         NDT(ray)-4size   &  0.208/$\mathbf{0.228}$(+10.1\%)  &  0.794/$\mathbf{0.834}$(+5.0\%) & 0.414/$\mathbf{0.509}$(+22.9\%) \\
         \bottomrule
        \end{tabular}
    }
    \caption{\textbf{Metrics of multi-scale EAG augmentation on Area2\_Bump.} Evaluation of 2$\times$ and 4$\times$ multi-scale augmentation using NDT and NDT(ray) decoders.}
    \label{tab:area2bump_multi_aug}
\end{table}

\subsection{Ablation studies}
Previous experiments typically set the exponential coefficient $\alpha$ to 1 in energy loss (Equation \ref{eq:energy_loss}). However, the energy score remains strictly proper for all $\alpha \in (0, 2)$. Since setting $\alpha < 1$ can lead to instability during early training due to unbounded gradients, we focus on the range $\alpha \in [1, 2)$. Notably, when $\alpha = 2$, the energy score is still proper while not strictly proper. The looser constraint ($\mathbb{E}_{p_{\theta}}[\mathbf{z}] = \mathbb{E}_q[\mathbf{z}_{\text{data}}]$) in this case is insufficient to effectively guide the model to accurately capture realistic neural dynamics. As demonstrated in Table \ref{tab:ablat_alpha}, models trained with $\alpha = 2$ exhibit clear deficiencies in generating realistic spiking activity, whereas models with $\alpha \in [1, 2)$ perform consistently well with only minor differences. Based on these results, we adopt $\alpha = 1$ as our default setting throughout all experiments.

\begin{table}[t]
    \centering
    \setlength{\tabcolsep}{1mm}
    \scalebox{0.76}{
        \begin{tabular}{c|c c c c}
        \toprule
        $\alpha$ & $D_{KL}$ psch & RMSE pairwise corr & RMSE mean isi & RMSE std isi \\
        \midrule
        1.0  & 0.0014 $\pm$ 2.0e-4 & 0.0024 $\pm$ 1.0e-5 & 0.024 $\pm$ 0.001  & 0.018 $\pm$ 0.0024 \\
        1.25 & 0.0026 $\pm$ 2.3e-4 & 0.0024 $\pm$ 1.0e-5 & 0.033 $\pm$ 0.007  & 0.018 $\pm$ 0.0007 \\
        1.5  & 0.0013 $\pm$ 9.8e-5 & 0.0025 $\pm$ 9.0e-6 & 0.022 $\pm$ 0.001  & 0.018 $\pm$ 0.0018 \\
        1.75 & 0.0017 $\pm$ 1.1e-4 & 0.0024 $\pm$ 1.2e-5 & 0.026 $\pm$ 0.001  & 0.019 $\pm$ 0.0007 \\
        2.0  &  0.0541 $\pm$ 7.8e-4 & 0.0028 $\pm$ 1.1e-5 & 0.051 $\pm$ 0.004  & 0.027 $\pm$ 0.0007 \\
        \bottomrule
        \end{tabular}
    }
    \caption{\textbf{Ablations on strict propriety.} Results of varying the exponential coefficient $\alpha$ in the energy loss, highlighting importance of strict propriety.}
    \label{tab:ablat_alpha}
\end{table}


\section{Conclusion}
We developed a novel Energy-based Autoregressive Generation (EAG) framework that resolves the fundamental trade-off between computational efficiency and high-fidelity modeling through latent energy-based learning. EAG achieves state-of-the-art generation quality with substantial computational efficiency improvements over existing methods, particularly diffusion-based approaches. Conditional generation applications demonstrate generalization to unseen behavioral contexts and improvement of motor BCI decoding accuracy using generated neural data. These results establish that neural population dynamics can be effectively modeled through direct energy-based generation in latent space, eliminating computationally expensive iterative sampling procedures. This framework provides a foundation for applications requiring both computational efficiency and biological realism in neural population modeling.


\section{Acknowledgments}

This work was supported by the Lingang Laboratory, Grant No.LGL-1987 and the Strategic
Priority Research Program of Chinese Academy of Sciences
(XDB1010302).

\bibliography{aaai2026}

\newpage
\input{supplementary}

\end{document}

%% file: supplementary.tex
\clearpage
\setcounter{figure}{0}
\renewcommand{\thefigure}{\arabic{figure}}

\setcounter{table}{0}
\renewcommand{\thetable}{\arabic{table}}

\captionsetup[figure]{labelformat=default, name=Figure}

\section*{Supplementary Materials}

\subsection{Contents}

\begin{itemize}
    \item S.1 EAG Architecture
    \item S.2 Hyperparameters and computational resources
    \item S.3 Datasets
    \item S.4 Baselines
    \item S.5 Supplementary figures on Lorenz dataset
    \item S.6 Supplementary figures on MC\_Maze dataset
    \item S.7 Supplementary figures on Area2\_Bump dataset
    \item S.8 Supplementary baseline comparison after spike-history augmented
    \item S.9 Supplementary efficiency-quality comparisons
    \item S.10 Supplementary figures on EAG generalization to unseen labels
    \item S.11 Supplementary tables for EAG-augmented BCI decoding
    \item S.12 Supplementary tables on ablation study
\end{itemize}

\vspace{2em}
\subsection{S.1\hspace{1em}EAG Architecture}

Here, we describe the detailed network components of autoencoder and energy transformer in EAG.

\subsubsection{AutoEncoder.}

The EAG autoencoder comprises encoder and decoder networks, each built from an input MLP, a stack of S4 blocks, and an output MLP. Spike trains are first projected into a learned embedding space by the input MLP, then processed through several encoder blocks to yield low-dimensional neural latents via the output MLP. The decoder mirrors this flow in reverse: it maps latents through its input MLP, evolves them across decoder blocks, and finally reconstructs firing rates with the output MLP.

Inside each block, a time-mixing S4 layer applies one univariate state-space filter per channel. A small MLP with GELU activations then mixes the channels together. We use bidirectional S4 layers so the model sees both past and future context. During training, we apply coordinated dropout by masking random time bins and computing the Poisson negative log-likelihood only on those bins, which prevents the network from simply memorizing the observed rates. Compared to LDNS, our decoder is deeper with more S4 blocks, matching the higher capacity of the energy transformer backbone.

\subsubsection{Energy Transformer.}

The energy transformer employs an energy-based autoregressive architecture to generate neural latents within the autoencoder’s learned latent space. The detailed generation process, including output generation guided by energy score, masked autoregressive strategy and classifier-free guidance, has been described in the Methods Section. Here we focus on the transformer’s encoder–decoder architecture and the preprocessing of different conditional labels.

The encoder adopts a Vision Transformer (ViT) architecture but, like Masked Autoencoder (MAE), it only processes the unmasked latent positions. Each visible latent position is linearly projected, enriched with positional embeddings, and passed through a series of transformer blocks, while masked positions are simply dropped. The decoder then reconstructs the full token set by combining the encoded visible positions with explicit mask tokens. All tokens receive positional embeddings and are jointly updated through another stack of transformer blocks, while the energy loss is applied only to the unmasked time points.

Conditional labels are incorporated by first projecting them into the model’s embedding space via a small MLP. The initial reach angle $\alpha_{\text{init}}$ is converted into its $(\cos\alpha_{\text{init}}, \sin\alpha_{\text{init}})$ representation, mapped to the transformer’s dimensionality, and concatenated with the neural latents along the temporal axis. Hand-velocity labels $(v_x, v_y)$ undergo the MLP projection and are appended as two additional channels alongside the neural inputs. This simple yet effective preprocessing allows the energy transformer to attend jointly to neural activity and behavioral context during latent generation.

\subsection{S.2\hspace{1em}Hyperparameters and computational resources}

\subsubsection{Hyperparameters.} Hyperparameter settings of autoencoder and energy transformer are shown in Table \ref{tab:append_ae_params} and Table \ref{tab:append_ET_params} under both conditional and unconditional generation. The first stage training settings follow the LDNS settings. The second stage training settings are shown in Table \ref{tab:append_train_params} under unconditional generation and conditional generation with an additional parameter cfg set to 4.

\begin{table}[htb]
    \centering
    \setlength{\tabcolsep}{1mm}
    \scalebox{0.95}{
        \begin{tabular}{l|ccc}
            \toprule
            \textbf{Parameter} & \textbf{Lorenz} & \textbf{MC\_Maze} & \textbf{Area2\_Bump} \\
            \midrule
            Encoder blocks & 4 & 4 & 4 \\
            Decoder blocks & 4 & 4 & 1 \\
            Linear layes & 4 & 4 & 4 \\
            Embed dim & 256 & 256 & 256 \\
            Num latents & 8 & 16 & 16 \\
            \bottomrule
        \end{tabular}
    }
    
    \caption{\textbf{Autoencoder parameters.} Detailed setting of autoencoder parameters under both unconditional generation and conditional generation.}
    \label{tab:append_ae_params}
\end{table}

\begin{table}[htb]
    \centering
    \setlength{\tabcolsep}{1mm}
    \scalebox{0.95}{
        \begin{tabular}{l|ccc}
            \toprule
            \textbf{Parameter} & \textbf{Lorenz} & \textbf{MC\_Maze} & \textbf{Area2\_Bump} \\
            \midrule
            \textbf{ViT Encoder} & & & \\
            Embed dim & 256 & 256 & 384 \\
            Depth & 4 & 4 & 6 \\
            Num heads & 4 & 4 & 6 \\
            \midrule
            \textbf{ViT Decoder} & & & \\
            Embed dim & 256 & 256 & 384 \\
            Depth & 4 & 4 & 6 \\
            Num heads & 4 & 4 & 6 \\
            \midrule
            \textbf{MLP Generator} & & & \\
            Depth & 6 & 6 & 6 \\
            Width & 768 & 768 & 768 \\
            Noise channels & 64 & 64 & 64 \\
            \bottomrule
        \end{tabular}
    }
    
    \caption{\textbf{Energy transformer parameters.} Detailed setting of ViT and MLP generator parameters in energy transformer under both unconditional generation and conditional generation.}
    \label{tab:append_ET_params}
\end{table}

\begin{table}[htb]
    \centering
    \setlength{\tabcolsep}{1mm}
    \scalebox{0.95}{
        \begin{tabular}{l|ccc}
            \toprule
            \textbf{Parameter} & \textbf{Lorenz} & \textbf{MC\_Maze} & \textbf{Area2\_Bump} \\
            \midrule
            Learning rate & 1e-4 & 1e-4 & 1e-4 \\
            Num epochs & 4000 & 4000 & 4000 \\
            Num warmup epochs & 100 & 100 & 100 \\
            Batch size & 512 & 256 & 256 \\
            Alpha & 1.0 & 1.0 & 1.0 \\
            Autoregression steps & 64 & 50 & 50 \\
            Train temperature & 1.0 & 1.0 & 1.0 \\
            Infer temperature & 0.7 & 0.7 & 0.7 \\
            \bottomrule
        \end{tabular}
    }

    \caption{\textbf{Training parameters.} Detailed setting of parameters for the second stage training under both unconditional generation and conditional generation.}
    \label{tab:append_train_params}
\end{table}

\subsubsection{Computational Resources.} 

All experiments are carried out on NVIDIA A40 GPUs with 40 GB of memory. We allocate two GPUs for the Lorenz and MC\_Maze datasets, and a single GPU for Area2\_Bump.

On the Lorenz dataset, we train the autoencoder for approximately 15 minutes, and train the energy transformer for about 6 hours. Generating roughly 7,000 samples takes around 40 seconds. In total, the GPU wall-clock time for both training and evaluation is about 6.5 hours.

For MC\_Maze, the autoencoder converges in about 10 minutes, and the energy transformer is trained for about 8 hours. Sampling $\sim$2,000 trials requires only 10 seconds, bringing the overall GPU wall-clock time to roughly 8.5 hours.

In the Area2\_Bump experiments, all computations run on a single GPU. The autoencoder completes training in 4 minutes, and the energy transformer finishes in 2.5 hours. Sampling \~300 trials takes about 5 seconds, for a total GPU utilization of under 3 hours.

\subsection{S.3\hspace{1em}Datasets}

We use two publicly available real-world datasets, both shared under open-access licenses. No new experimental data were collected in this study. The parameters of three dataset are show in Table \ref{tab:append_dataset}.

\begin{table}[htb]
    \centering
    \setlength{\tabcolsep}{1mm}
    \scalebox{0.9}{
        \begin{tabular}{l|ccc}
        \toprule
        \textbf{Parameter} & \textbf{Lorenz} & \textbf{MC\_Maze} & \textbf{Area2\_Bump} \\
        \midrule
        Num training trials & 7000 & 2008 & 318 \\
        Trial length (bins) & 256 & 140 & 140 \\
        Data channels (neurons) & 128 & 182 & 65 \\
        \bottomrule
        \end{tabular}
    }
    \caption{\textbf{Dataset details.} Number of training trials, trial length (bins) and number of neurons across the three datasets.}
    \label{tab:append_dataset}
\end{table}

\textbf{MC\_Maze Dataset}. The MC\_Maze dataset \cite{churchland2022mc_maze, pei2021neural} is available on the DANDI Archive (https://dandiarchive.org/dandiset/000128, ID: 000128) under a CC-BY-4.0 license. It was recorded from the premotor and primary motor cortices (182 units) of a monkey performing a delayed center-out reaching task with barriers. The dataset primarily captures autonomous, motor-related neural dynamics.

\textbf{Area2\_Bump Dataset}. The Area2\_Bump dataset \cite{pei2021neural,chowdhury2020area} is available on the DANDI Archive (https://dandiarchive.org
/dandiset/000127, ID: 000127). It was recorded from the somatosensory cortex (65 units) during a reaching task with external bump perturbations. This dataset reflects both predictable and unpredictable input-driven neural responses.

\subsection{S.4\hspace{1em}Baselines}

\subsubsection{Targeted Neural Dynamical Modeling (TNDM).} TNDM \cite{hurwitz2021targeted} is a nonlinear state-space model implemented as a sequential variational autoencoder that simultaneously models neural activity and external behavioral variables. A key feature of TNDM is its disentangled latent space, separating task-relevant and intrinsic neural dynamics into behavior-relevant and behavior-independent subspaces. The behaviorally relevant dynamics are decoded to reconstruct behavior via a flexible linear decoder, while both components are combined to reconstruct neural activity through a linear decoder without time lag.

In our experiments, we train a TNDM model following the architecture and hyperparameters from the original implementation. We used 64-dimensional latent dynamics for both subspaces, yielding 10 total latent factors $\mathbf{z}$ (5 behavior-relevant $\mathbf{z_r}$ and 5 behavior-independent $\mathbf{z_i}$). To perform unconditional generation, we sample initial states from a standard Gaussian prior $\mathcal{N}(0,I)$ and use the generative process defined in LFADS. Across all population-level and single-neuron metrics, our EAG model consistently outperforms TNDM.

\subsubsection{Poisson-identifiable VAE (pi-VAE).} Poisson-identifiable VAE builds on the insight that conditioning the latent variable $\mathbf{z}$ on an auxiliary variable $\mathbf{u}$ restores identifiability. Based on that, pi-VAE achieves latent space identifiability by conditioning the latent variables on external behavioral variables. It models neural spiking as Poisson-distributed observations but treats each time bin independently, lacking temporal dynamics.

In our experiments, we train a pi-VAE model following the architecture and hyperparameters from the original implementation. A flexible normalizing-flow decoder (the General Incompressible-Flow Network) maps four latent dimensions (two relevant, two independent) back to observed firing rates. For evaluation, we use 5ms time bins instead of 50ms used in pi-VAE. Besides, we use more complex behavior labels, comprising cosine and sine of initial and final reach angles. As expected, the lack of temporal dynamics limits pi-VAE’s ability to produce realistic spike trains, and it underperforms our EAG model on all spike-statistic benchmarks.

\subsubsection{Latent Factor Analysis via Dynamical Systems (LFADS).} LFADS is a sequential variational autoencoder that infers low-dimensional neural dynamics by first using an encoder RNN to map observed spike trains into a posterior over the generator’s initial state. At each time step, an optional controller RNN combines the encoded neural activity with the previous latent factors to produce time-varying inputs that drive the generator. The generator RNN evolves its hidden state, and its internal states are mapped to low-dimensional latent factors and single-neuron Poisson firing rates through affine transformations. AutoLFADS extends this framework by applying Population-Based Training (PBT) to automatically search and optimize hyperparameters, reducing manual tuning and improving model stability.

We train the PyTorch AutoLFADS implementation on 5 ms–binned monkey reach data using the original hyperparameter ranges but omit the controller to simplify generation. Because center-out reaching can be well modeled as autonomous \cite{churchland2012neural}, and LFADS already performs well without controller inputs \cite{pandarinath2018inferring}, this omission does not impair its performance. We sample initial generator states from a standard Gaussian prior, evolve dynamics via the generator RNN, and draw spikes from the resulting Poisson rates. Our EAG model outperforms both AutoLFADS and its spike-history–augmented variant (AutoLFADSsh) across all metrics.




\setcounter{figure}{0}
\renewcommand{\thefigure}{S\arabic{figure}}

\setcounter{table}{0}
\renewcommand{\thetable}{S\arabic{table}}

\captionsetup[figure]{labelformat=default, name=Figure}

\clearpage
\onecolumn
\subsection{S.5\hspace{1em}Supplementary figures on Lorenz dataset}
\vspace{2em}

\begin{figure}[h]
    \centering

    \begin{overpic}[width=0.8\textwidth]{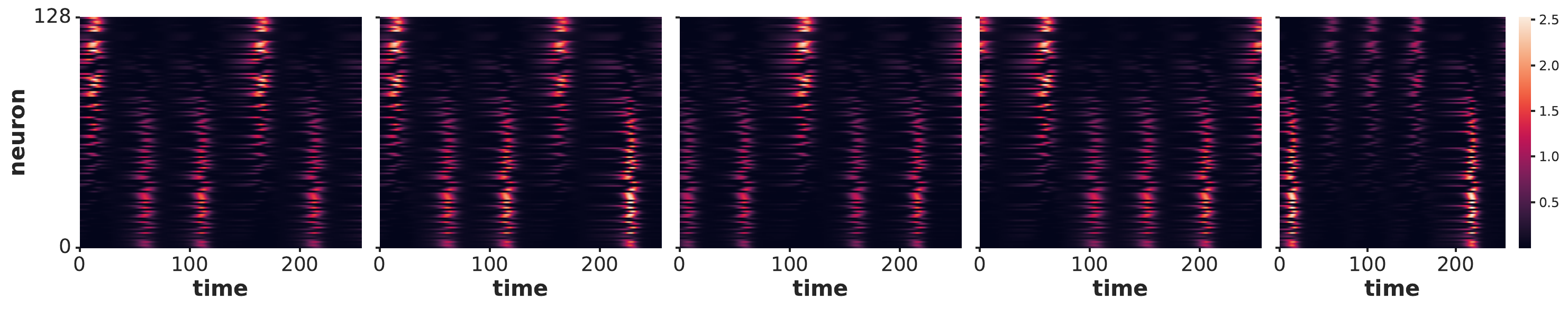}
        \put(-5,20){\textbf{(a)}}
    \end{overpic}

    \vspace{1em}

    \begin{overpic}[width=0.8\textwidth]{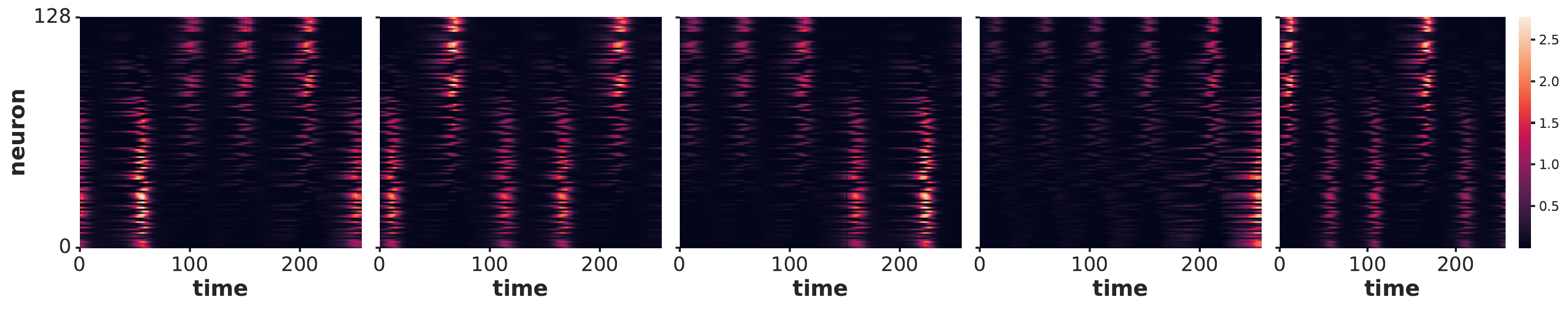}
        \put(-5,20){\textbf{(b)}}
    \end{overpic}

    \caption{\textbf{Real rates and sampled rates.} (a) Real firing rates and (b) sampled rates on the Lorenz dataset. Five randomly selected examples are shown for each. The sampled rates closely resemble the real ones in their temporal structure, making it visually indistinguishable to tell them apart.}
    \label{fig:lorenz_rates}
\end{figure}

\begin{figure}[h]
    \centering

    \begin{overpic}[width=0.8\textwidth]{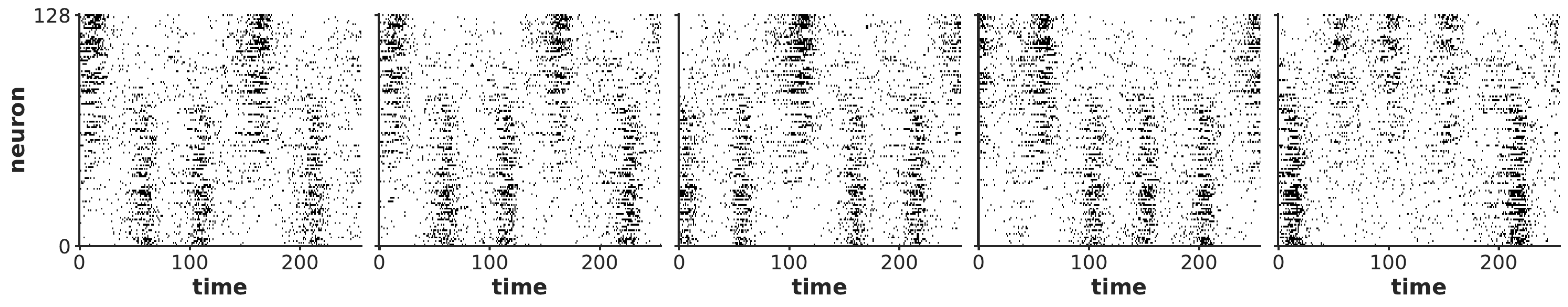}
        \put(-5,20){\textbf{(a)}}
    \end{overpic}

    \vspace{1em}

    \begin{overpic}[width=0.8\textwidth]{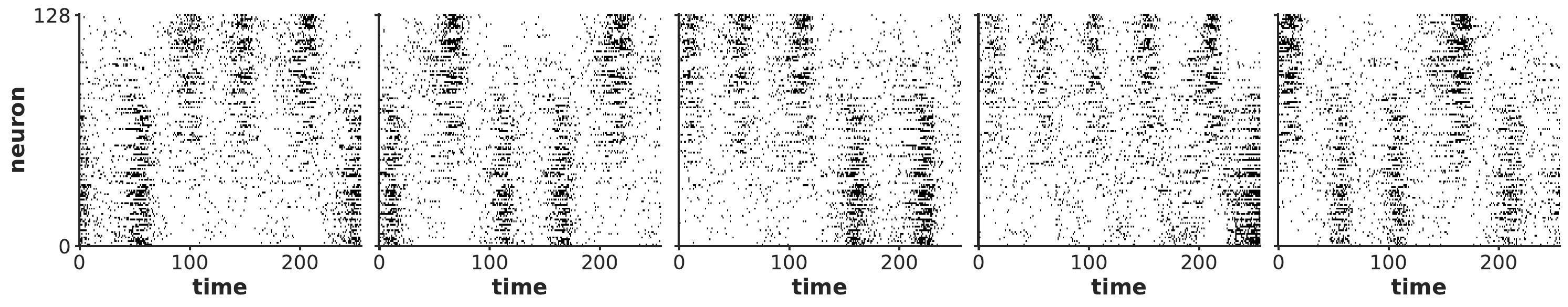}
        \put(-5,20){\textbf{(b)}}
    \end{overpic}

    \caption{\textbf{Real spikes and sampled spikes.} (a) Real spike trains and (b) sampled spike trains on the Lorenz dataset. The ten examples are corresponded to the rates examples above. The sampled spikes exhibit similar temporal patterns to the real ones, making it visually indistinguishable.}
    \label{fig:lorenz_spikes}
\end{figure}

To provide a more intuitive supplement to the main text, we show additional examples of EAG‐generated firing rates and spike trains on the Lorenz dataset, as illustrated in the Figure \ref{fig:lorenz_rates} and Figure \ref{fig:lorenz_spikes}. Both the sampled rates and spikes faithfully reproduce the firing patterns and temporal dynamics of the real data, making it difficult to distinguish the generated outputs from the real ones by visual inspection. We also present the pairwise correlations of firing rates reconstructed by the autoencoder and those generated by EAG. As shown in Figure \ref{fig:lorenz_corr}, both the autoencoder-reconstructed and EAG-generated rates closely align with the ground-truth pairwise correlations, indicating that the population-level structure is accurately preserved.

\vspace{2em}

\vspace{2em}


\clearpage
\vspace{4em}

\begin{center}
    \includegraphics[width=0.8\textwidth]{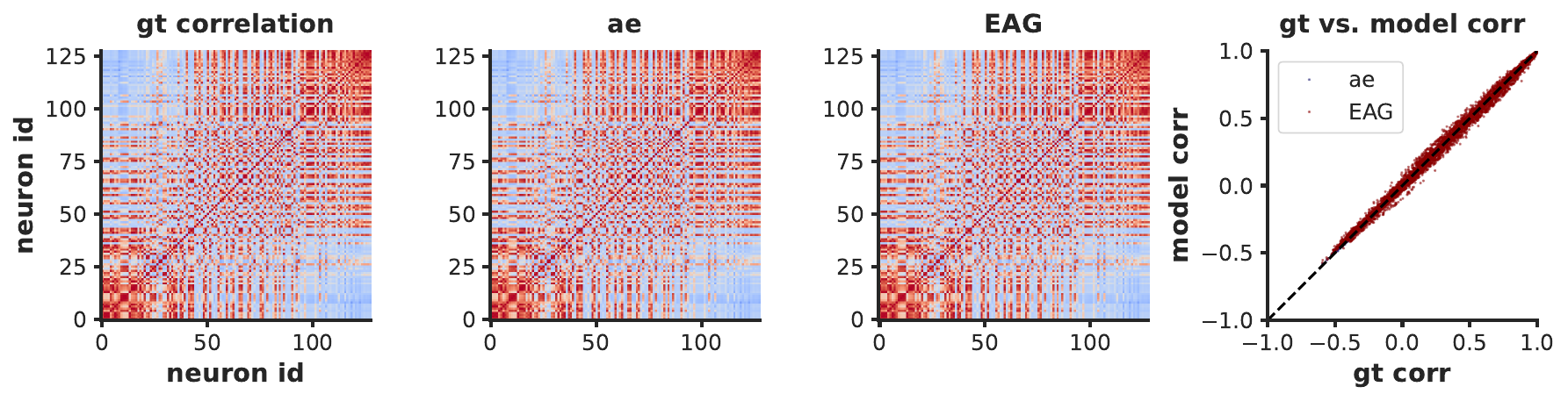}
    \captionof{figure}{\textbf{Correlation matrix of the Lorenz dataset.} The AE-reconstructed and EAG-generated firing rates closely match the ground-truth pairwise correlations, indicating accurate preservation of the underlying population-level structure.}
    \label{fig:lorenz_corr}
\end{center}

\clearpage
\subsection{S.6\hspace{1em}Supplementary figures on MC\_Maze dataset}
\vspace{2em}

\begin{figure}[h]
    \centering

    \begin{overpic}[width=0.8\textwidth]{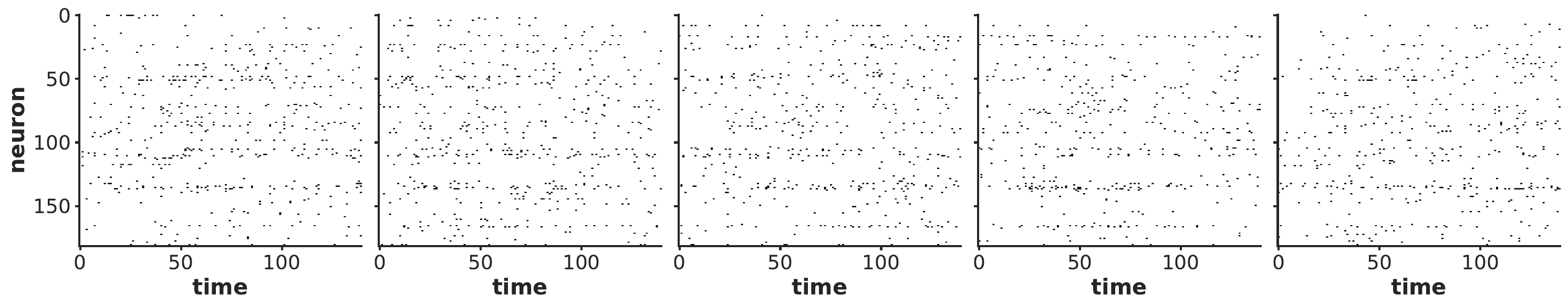}
        \put(-5,20){\textbf{(a)}}
    \end{overpic}

    \vspace{1em}

    \begin{overpic}[width=0.8\textwidth]{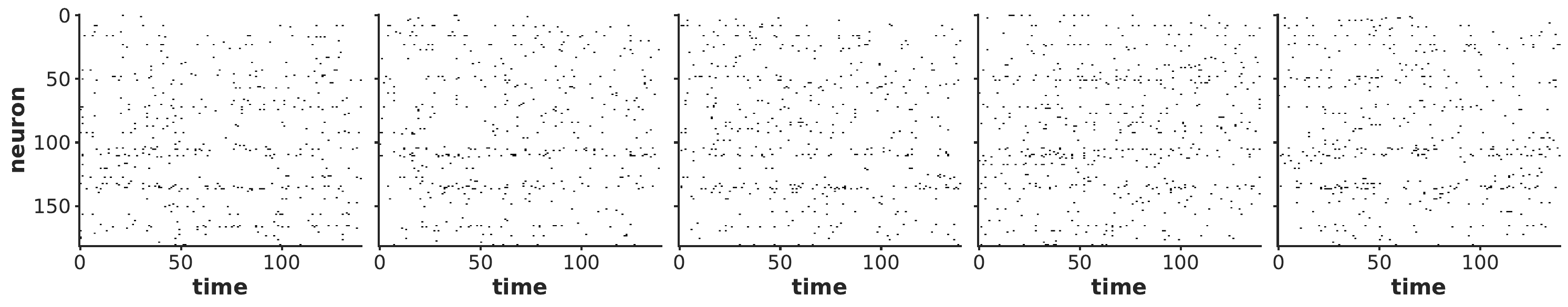}
        \put(-5,20){\textbf{(b)}}
    \end{overpic}

    \caption{\textbf{Real spikes and sampled spikes.} (a) Real spike trains and (b) sampled spike trains on the MC\_Maze dataset. Five randomly selected examples are shown for each. The sampled spikes closely resemble the real ones in their temporal structure, making it visually indistinguishable to tell them apart.}
    \label{fig:mcmaze_spikes}
\end{figure}

\begin{figure}[htb]
    \centering
    \includegraphics[scale=0.4]{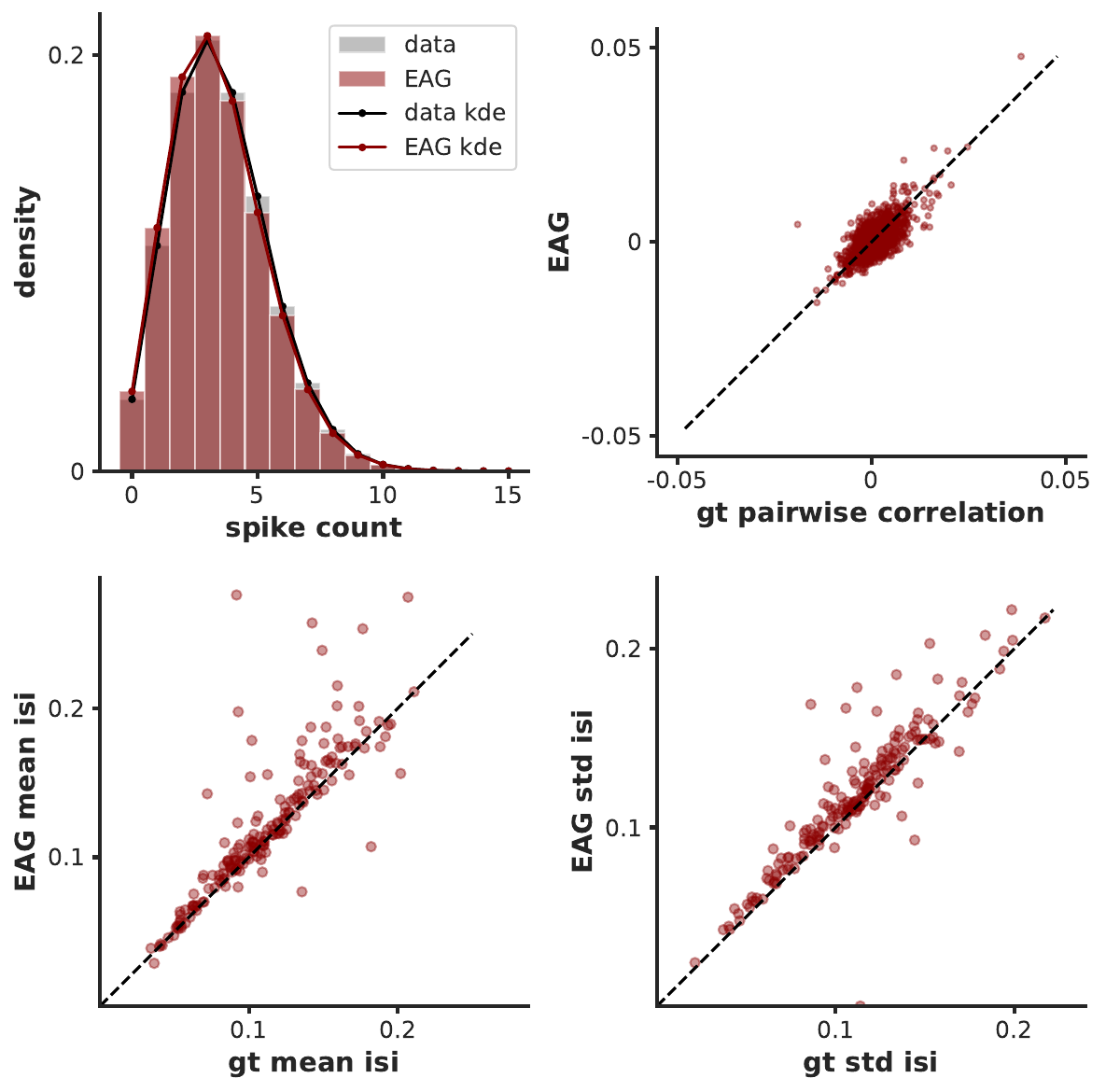}
    \caption{\textbf{Metrics on MC\_Maze Dataset.} This figure visualizes results between unconditional sampled rates and real rates on four quantitative metrics on MC\_Maze dataset. Sampled rates closely match real rates, indicating that EAG captures both population-level dynamics and single-neuron firing patterns.}
    \label{fig:mcmaze_metrics}
\end{figure}

As a complement to the main text presenting unconditional generation results on the MC\_Maze dataset, we visualized the sampled spike trains, as shown in the Figure \ref{fig:mcmaze_spikes}. Figure \ref{fig:mcmaze_spikes} demonstrates that the sampled spike trains are visually indistinguishable from the real spike trains. Furthermore, we visualized the unconditional generation results across four evaluation metrics, as illustrated in Figure \ref{fig:mcmaze_metrics}. These results reveal that the sampled spikes closely match the statistical properties of the real spikes, indicating that EAG effectively captures both the population-level co-activity firing patterns and the single-neuron firing regularities.

\vspace{2em}

\clearpage
\subsection{S.7\hspace{1em}Supplementary figures on Area2\_bump dataset}
\vspace{2em}

\begin{figure}[h]
    \centering

    \begin{overpic}[width=0.8\textwidth]{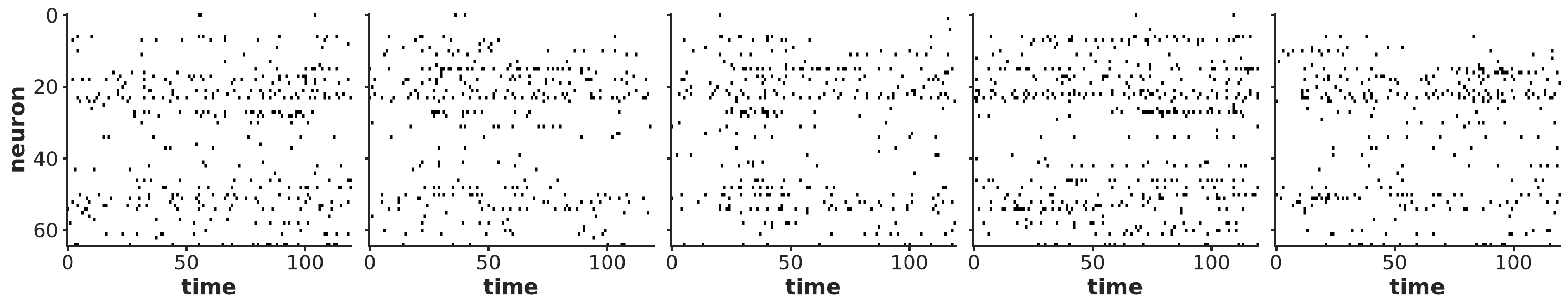}
        \put(-5,20){\textbf{(a)}}
    \end{overpic}

    \vspace{1em}

    \begin{overpic}[width=0.8\textwidth]{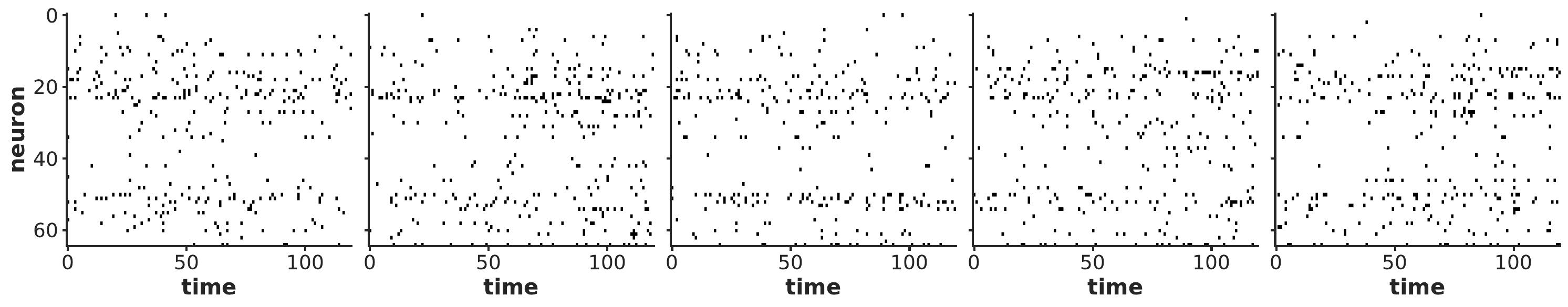}
        \put(-5,20){\textbf{(b)}}
    \end{overpic}

    \caption{\textbf{Real spikes and sampled spikes.} (a) Real firing rates and (b) sampled rates on the Area2\_Bump dataset. Five randomly selected examples are shown for each. The sampled rates closely resemble the real ones in their temporal structure, making it visually indistinguishable to tell them apart.}
    \label{fig:area2bump_spikes}
\end{figure}

\begin{figure}[htb]
    \centering
    \includegraphics[scale=0.4]{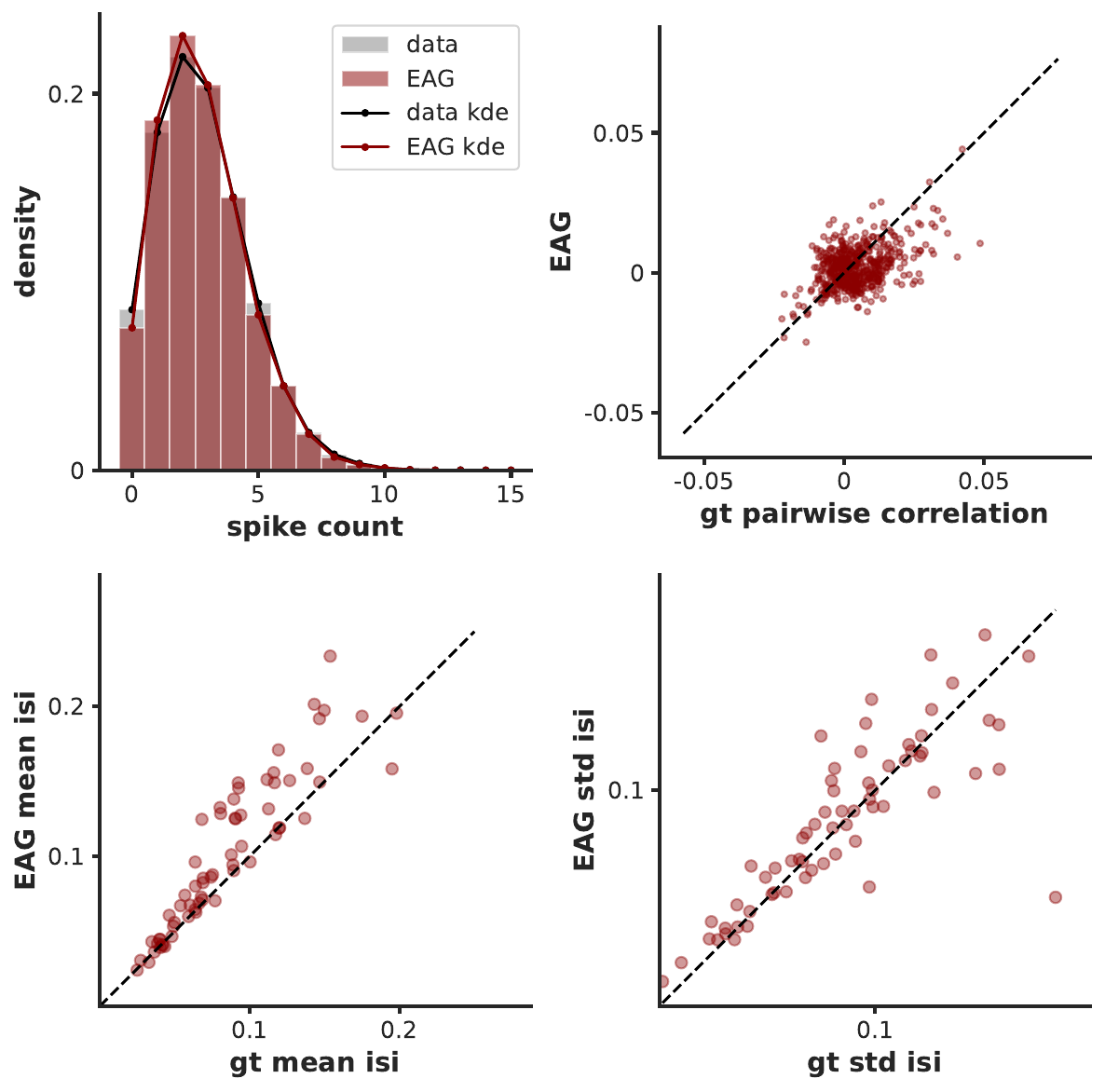}
    \caption{\textbf{Metrics on Area2\_Bump Dataset.} This figure visualizes results between unconditional sampled rates and real rates on four quantitative metrics on Area2\_Bump dataset. Sampled rates closely match real rates, indicating that EAG captures both population-level dynamics and single-neuron firing patterns.}
    \label{fig:area2bump_metrics}
\end{figure}

To further support the results of unconditional generation on the small-scale Area2\_Bump dataset presented in the main text, we provide a visualization of the sampled spike trains in Figure \ref{fig:area2bump_spikes}. As shown, the generated spike trains exhibit a high degree of visual similarity to the ground-truth recordings, making them nearly indistinguishable by eye. In addition, we visualize the performance of unconditional generation on four quantitative metrics, presented in Figure \ref{fig:area2bump_metrics}. The results demonstrate a close alignment between the statistical characteristics of the generated and real spikes, highlighting EAG’s capability to accurately model both population-level synchronous activity and the temporal structure of individual neuron firing.

\vspace{2em}

\clearpage
\subsection{S.8\hspace{1em}Supplementary baseline comparison after spike-history augmented}
\vspace{2em}
\subsubsection{Spike History Approach}
The spike history approach augments the observation model by incorporating each neuron's own spiking history, capturing neuron-specific dynamics like refractory periods or burstiness. It enhances the realism of generated neural data without changing the underlying latent dynamics. As a result, the model better reproduces single-neuron autocorrelation patterns, making synthetic activity more biologically plausible. However, this is more of an auxiliary statistical correction than a principled modeling improvement, as it does not modify the underlying latent dynamics. As shown in Table \ref{tab:mcmaze_metrics_sh} and Table \ref{tab:area2bump_metrics_sh}, the results with EAG suggest that such post-hoc adjustments may compensate for limited model capacity, but are not a substitute for fundamentally stronger generative models.

\vspace{2em}

\begin{table*}[h]
  \centering
  \begin{tabular}{l|cccc}
    \toprule
    Method & $D_{KL}$ psch & RMSE pairwise corr & RMSE mean isi & RMSE std isi \\
    \midrule
    TNDM & 0.0028 $\pm$ 6.0e-5 & 0.0027 $\pm$ 1.17e-5 & 0.057 $\pm$ 0.004 & 0.029 $\pm$ 0.001 \\
    pi-VAE & 0.0063 $\pm$ 2.0e-4 & 0.0031 $\pm$ 1.08e-5 & 0.064 $\pm$ 0.002 & 0.034 $\pm$ 0.001 \\
    Transformer(plain) & 0.0036 $\pm$ 2.0e-4 & 0.0029 $\pm$ 1.8e-5 & 0.051 $\pm$ 0.005 & 0.032 $\pm$ 0.001 \\
    AutoLFADS & 0.0040 $\pm$ 2.2e-4 & 0.0026 $\pm$ 1.25e-5 & 0.039 $\pm$ 0.003 & 0.029 $\pm$ 0.001 \\
    LDNS & 0.0039 $\pm$ 3.0e-4 & 0.0025 $\pm$ 1.1e-4 & 0.037 $\pm$ 0.001 & 0.023 $\pm$ 0.0001 \\
    \midrule
    AutoLFADSsh & 0.0036 $\pm$ 2.1e-4 & 0.0026 $\pm$ 1.8e-5 & 0.034 $\pm$ 0.002 & 0.023 $\pm$ 0.0001 \\
    LDNSsh & 0.0016 $\pm$ 6.2e-4 & 0.0025 $\pm$ 1.07e-5 & 0.024 $\pm$ 0.002 & 0.023 $\pm$ 0.0001 \\
    \midrule
    EAG & $\mathbf{0.0014}$ $\pm$ 2e-4 & $\mathbf{0.0024}$ $\pm$ 1e-5 & $\mathbf{0.0024}$ $\pm$ 0.001 & $\mathbf{0.018}$ $\pm$ 0.0024 \\
    \bottomrule
  \end{tabular}
  \caption{\textbf{Metrics comparison on MC\_Maze with spike history augmentation.} Even after augmenting baseline models (AutoLFADS and LDNS) with the spike history approach, EAG consistently outperforms them, indicating its more effective modeling without relying on such auxiliary enhancements.}
  \label{tab:mcmaze_metrics_sh}
\end{table*}

\vspace{2em}
\begin{table*}[h]
  \centering
  \begin{tabular}{l|cccc}
    \toprule
    Method & $D_{KL}$ psch & RMSE pairwise corr & RMSE mean isi & RMSE std isi \\
    \midrule
    TNDM      & 0.0027 $\pm$ 2.9e-4 & 0.0077 $\pm$ 1.0e-4 & 0.049 $\pm$ 0.009 & 0.029 $\pm$ 0.003 \\
    pi-VAE    & 0.0067 $\pm$ 4.2e-4 & 0.0088 $\pm$ 7.9e-5 & 0.050 $\pm$ 0.007 & 0.029 $\pm$ 0.004 \\
    AutoLFADS & 0.0032 $\pm$ 3.2e-4 & 0.0081 $\pm$ 1.2e-5 & 0.048 $\pm$ 0.003 & 0.031 $\pm$ 0.006\\
    LDNS      & 0.0020 $\pm$ 1.2e-4 & 0.0076 $\pm$ 1.4e-5 & 0.050 $\pm$ 0.002 & 0.034 $\pm$ 0.002 \\
    \midrule
    AutoLFADSsh & 0.0028 $\pm$ 1.6e-4 & 0.0079 $\pm$ 2.2e-5 & 0.043 $\pm$ 0.005 & 0.034 $\pm$ 0.003\\
    LDNSsh      & 0.0020 $\pm$ 8.0e-4 & 0.0076 $\pm$ 1.7e-5 & 0.037 $\pm$ 0.005 & 0.036 $\pm$ 0.002 \\
    \midrule
    EAG & $\mathbf{0.0018}$ $\pm$ 1.6e-4 & \textbf{0.0075} $\pm$ \textbf{9.1e-5} & $\mathbf{0.035}$ $\pm$ 0.004 & $\mathbf{0.025}$ $\pm$ 0.003 \\
    \bottomrule
  \end{tabular}
  \caption{\textbf{Metrics comparison on Area2\_Bump with spike history augmentation.} The spike history approach yields limited improvement on this smaller-scale dataset, whereas EAG consistently maintains the best performance.}
  \label{tab:area2bump_metrics_sh}
\end{table*}

\clearpage
\subsection{S.9\hspace{1em}Supplementary efficiency-quality comparisons}
\vspace{2em}

\begin{figure}[h]
    \centering
    \includegraphics[scale=0.4]{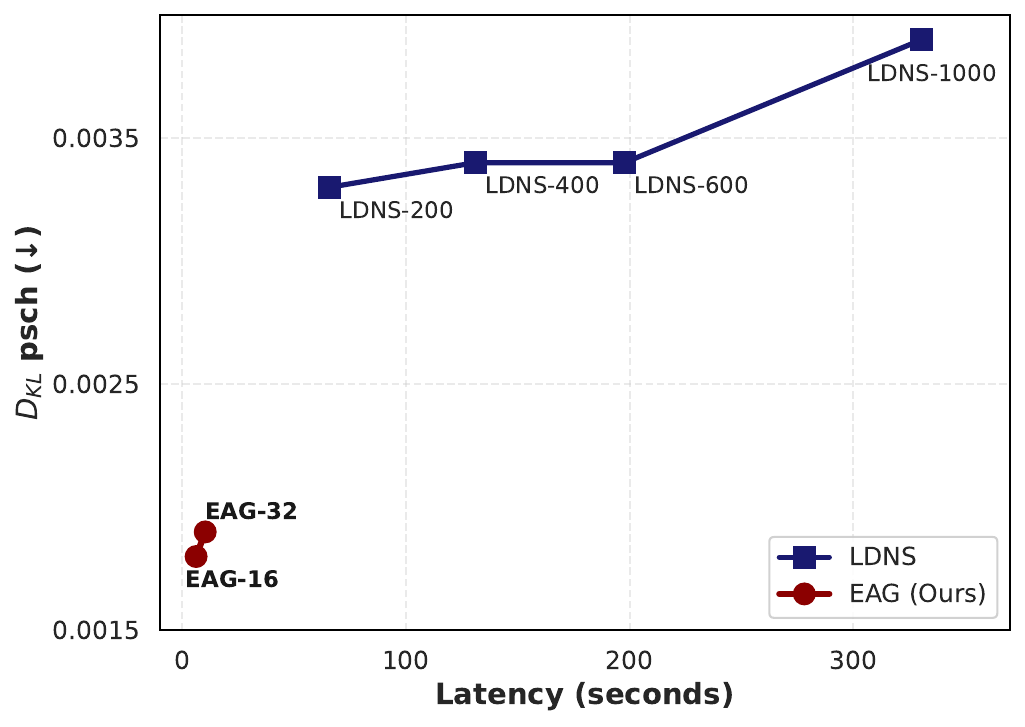}
    \caption{\textbf{The latency/quality trade-off for EAG and LDNS}. We vary number of diffusion steps (200, 400, 600, 1000) of LDNS and number of autoregressive steps (16, 32) of EAG. EAG-32 achieves a 96.9\% reduction in latency, and a 51.3\% improvement on $D_{KL}$ psch compared to LDNS-1000}
    \label{fig:eff_psch}
\end{figure}

While the main text presents the relationship between sample latency and RMSE of mean inter-spike interval (single-neuron metric), here we provide a complementary analysis focusing on the relationship between sample latency and the population-level statistic $D_{\mathrm{KL}}$ psch, as shown in Figure \ref{fig:eff_psch}. The results indicate that both EAG-16 and EAG-32 consistently achieve lower sample latency and reduced $D_{\mathrm{KL}}$ psch compared to all variants of the LDNS baseline. For instance, EAG-32 achieves a 96.9\% reduction in latency and a 51.3\% improvement in $D_{\mathrm{KL}}$ psch relative to LDNS-1000. Detailed numerical results are reported in Table \ref{tab:EAG_efficiency}. These findings highlight the substantial advantage of EAG in sampling efficiency while maintaining high-quality generation performance.

\vspace{3em}

\vspace{3em}
\begin{table*}[h]
    \centering
    \begin{tabular}{l| r r r r r}
    \toprule
      Method   & $D_{KL}$ psch & RMSE pairwise corr & RMSE mean isi & RMSE std isi & Latency(sec) \\
      \midrule
      LDNS-200   &  0.0033$\pm$6.1e-5  &  0.0027$\pm$1.2e-5 & 0.049 $\pm$ 0.002  &  0.026$\pm$0.001  & 65.77  \\
      LDNS-400   &  0.0034$\pm$4.7e-4  &  0.0025$\pm$6.0e-6 & 0.046 $\pm$ 0.002  &  0.024$\pm$0.001  & 130.93  \\
      LDNS-600   &  0.0034$\pm$2.8e-4  &  0.0025$\pm$1.8e-5 & 0.039 $\pm$ 0.002  &  0.024$\pm$0.001  & 197.73  \\
      LDNS-1000   &  0.0039$\pm$3.0e-4  &  0.0025$\pm$1.1e-4 & 0.037 $\pm$ 0.001  &  0.023$\pm$0.0001  & 330.64  \\
      \midrule
      EAG-16   &  0.0018$\pm$7.3e-5  & 0.0025 $\pm$ 1.9e-5  & 0.031 $\pm$ 0.002  & 0.021 $\pm$ 0.002    & 6.22  \\
      EAG-32   &   0.0019$\pm$1.5e-4  & 0.0024 $\pm$ 2.3e-5  & 0.025 $\pm$ 0.002  & 0.021 $\pm$ 1e-5    & 10.29  \\
    \bottomrule
    \end{tabular}
    \caption{\textbf{Efficiency comparison.} Detailed comparison of quality and latency across four metrics, evaluating both population-level dynamics and single-neuron firing patterns as well as their associated latencies.}
    \label{tab:EAG_efficiency}
\end{table*}

\clearpage 

\vspace{2em}
To investigate how the generation efficiency of EAG scales with the number of neurons and generation duration, we conducted a scaling analysis. We examined changes in runtime and memory usage as the number of neurons increased from 50 to 500 in increments of 50. Fortunately, since generation occurs in a low-dimensional latent space, the only additional cost comes from reconstructing the latent representation back into neuron space, which has a negligible impact on efficiency. In our tests, generating 64 samples of 1-second duration required 2.11–2.32 s of runtime, with memory usage varying by only 0.21 ± 0.02 MB. On the other hand, both runtime and memory usage increase linearly with generation duration. For a setup with 200 neurons and a 5 ms time bin, we measured the time and memory required to generate 64 samples at varying durations. Detailed numerical results are reported in Table \ref{tab:EAG_efficiency_scaling_analysis}. Since typical trial lengths are less than 2 s, the observed efficiency overhead is acceptable.


\vspace{2em}
\begin{table*}[h]
    \centering
    \begin{tabular}{c r r r r r}
    \toprule
      Duration  & Timebins  & Runtime &  Memory Usage \\
      \midrule
      0.5s   &   100   &  1.22s  &  0.11MB  \\
      1.0s   &   200   &  2.18s  &  0.20MB  \\
      1.5s   &   300   &  3.21s  &  0.30MB  \\
      2.5s   &   500   &  5.31s  &  0.51MB  \\
      5.0s   &   1000   &  12.27s  &  0.99MB  \\
    \bottomrule
    \end{tabular}
    \caption{\textbf{Scaling analysis.} Detailed results of scaling analysis on generation duration. Both runtime and memory usage increase linearly with generation duration. Since typical trial lengths are less than 2 s, the observed efficiency overhead is acceptable}
    \label{tab:EAG_efficiency_scaling_analysis}
\end{table*}

\clearpage
\subsection{S.10\hspace{1em}Supplementary figures on EAG generalization to unseen labels}

\vspace{2em}
\subsubsection{Closed-loop Validation.} 
\begin{figure}[h]
\centering
\includegraphics[scale=0.7]{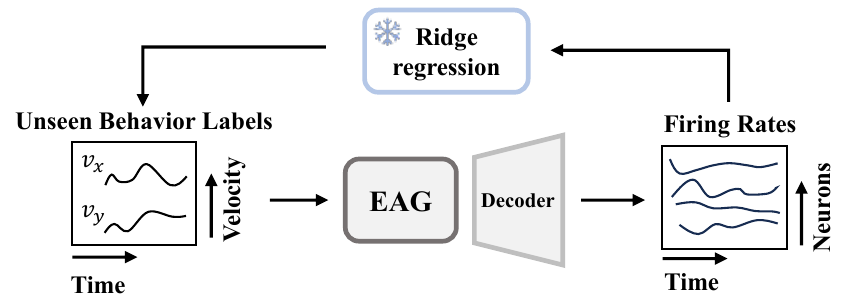}
\caption{\textbf{Closed-loop validation.} Schematic diagram of the closed-loop validation procedure.}
\label{fig:close_loop}
\end{figure}

A natural question in conditional generation is whether the generated firing rates truly correspond to the intended behavioral labels. To address this, we train a ridge regression model to predict behavior (e.g. velocity) from real firing rates (Figure \ref{fig:close_loop}), which is then used to decode behaviors from sampled rates. If the decoded behaviors match the conditioning behavioral labels, it validates that the generated activity is associated with the target behaviors.

\vspace{1em}

\vspace{1em}
\subsubsection{Conditional Generation on Angle Labels.} 

\begin{figure}[h]
    \centering

    \begin{overpic}[width=0.95\textwidth]{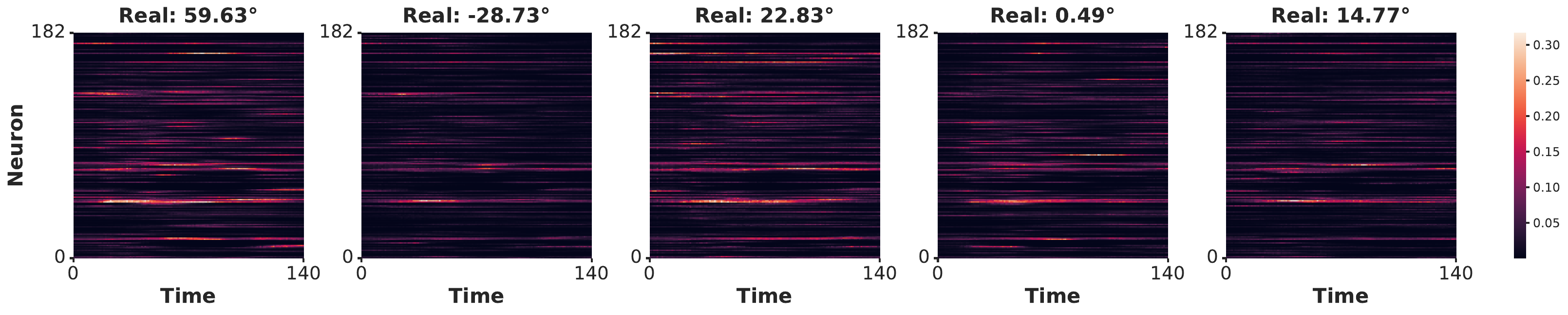}
        \put(-5,20){\textbf{(a)}}
    \end{overpic}

    \vspace{1em}

    \begin{overpic}[width=0.95\textwidth]{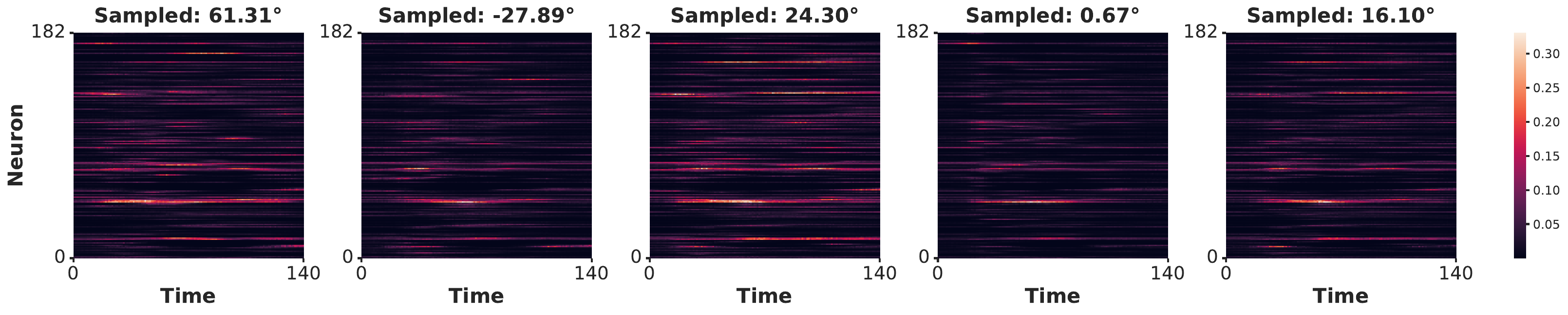}
        \put(-5,20){\textbf{(b)}}
    \end{overpic}

    \caption{\textbf{Real and sampled firing rates on unseen angle contexts.} Panel (b) displays EAG-generated firing rates for five randomly selected, unseen angle labels. Panel (a) shows the real firing rates for the closest available angle labels (since the unseen labels in (b) are not present in the dataset). The sampled rates closely match the firing patterns of the real rates while preserving trial-to-trial variability.}
    \label{fig:supp_angcond_rates}
\end{figure}

As a supplement to the conditional generation results based on angle labels presented in the main text, we provide additional examples of conditionally generated firing rates, as shown in Figure \ref{fig:supp_angcond_rates}. We randomly selected five sampled firing rate sequences, along with their corresponding angle labels, as illustrated in Figure \ref{fig:supp_angcond_rates}b. Since these angles are not present in the original dataset, we identified the closest real angles for comparison; the firing rates corresponding to these real angles are shown in Figure \ref{fig:supp_angcond_rates}a. Despite slight differences between the generated and real angles, the comparison reveals that EAG successfully generalizes to previously unseen angles, producing firing patterns highly consistent with those observed under real conditions. Notably, careful inspection confirms that the sampled rates are not mere replicas of real rates, but instead preserve trial-to-trial variability, which is typically not captured by prior predictive models.

\vspace{1em}

\clearpage
\vspace{4em}
\subsubsection{Conditional Generation on Velocity Labels.}
\begin{figure}[h]
    \centering

    \begin{overpic}[width=0.95\textwidth]{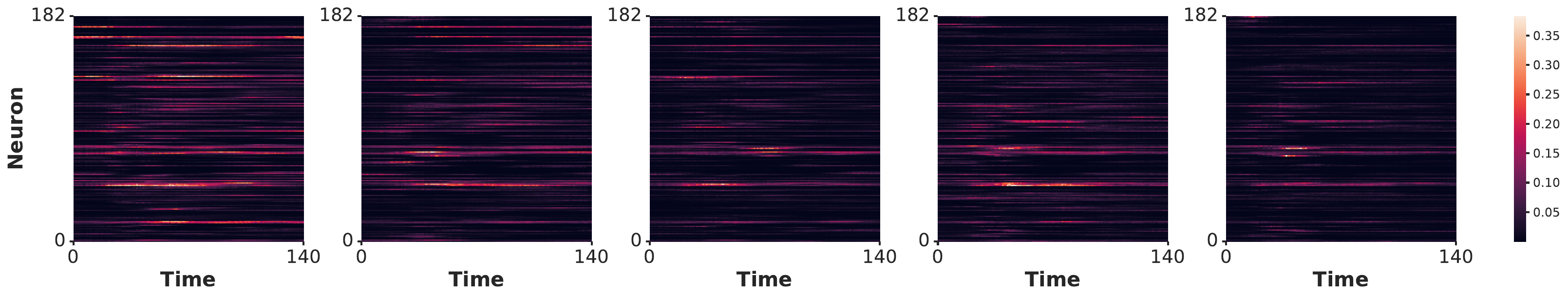}
        \put(-5,20){\textbf{(a)}}
    \end{overpic}

    \begin{overpic}[width=0.95\textwidth]{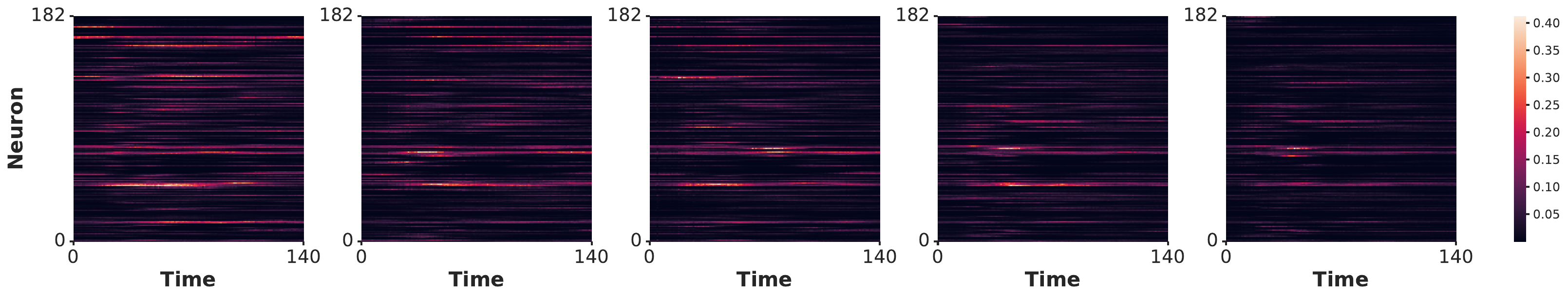}
        \put(-5,20){\textbf{(b)}}
    \end{overpic}

    \begin{overpic}[width=0.95\textwidth]{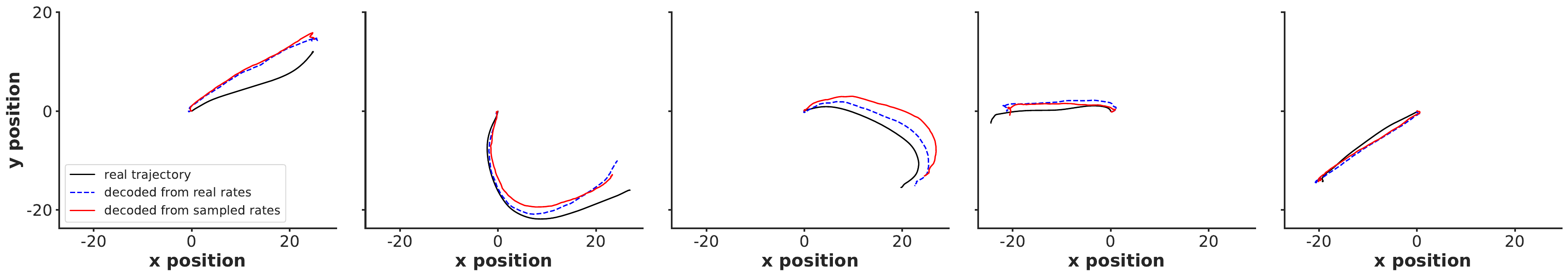}
        \put(-5,18){\textbf{(c)}}
    \end{overpic}

    \caption{\textbf{Real and sampled firing rates on unseen velocity contexts.} Panel (a) shows the true firing rates for five velocity labels that were held out from the training set, and panel (b) displays the corresponding EAG-generated firing rates for those same unseen labels. Sampled rates in panel (b) match well to real rates in panel (a) due to fine-grained label condition. Panel (c) shows the movement trajectories corresponding to these unseen velocity labels, with decoded trajectories from real rates and sampled rates aligning closely.}
    \label{fig:supp_velcond_rates}
\end{figure}

To further validate the conditional generation capabilities of EAG, we present additional examples conditioned on velocity labels that were held out during training, as shown in Figure \ref{fig:supp_velcond_rates}. Specifically, Figure \ref{fig:supp_velcond_rates}a displays the ground-truth firing rates for five unseen velocity conditions, while Figure \ref{fig:supp_velcond_rates}b shows the corresponding EAG-generated firing rates under the same label conditions. The close similarity between the sampled and real firing patterns demonstrates that EAG can accurately generalize to novel velocity contexts, benefiting from the fine-grained conditioning provided by continuous velocity labels. In addition, Figure \ref{fig:supp_velcond_rates}c compares the movement trajectories decoded from both real and sampled firing rates. The near alignment between these decoded trajectories further supports the consistency and fidelity of the generated neural activity, highlighting EAG’s ability to generalize to hypothetical movement while maintaining trial-to-trial variability.

\vspace{2em}




\clearpage
\subsection{S.11\hspace{1em}Supplementary tables for EAG-augmented BCI decoding}

\vspace{1em}

As summarized in Table \ref{tab:area2bump_aug}, results on the Area2\_Bump dataset show that EAG-based data augmentation consistently enhances decoding performance, with more substantial improvements observed in models of higher complexity. Among all models, the transformer-based Neural Data Transformer (NDT) demonstrates the largest performance gain, achieving up to a 42.5\% improvement in the co-smoothing bps metric. While most models benefit notably from EAG augmentation, the NDT with ray-based hyperparameter optimization exhibits no improvement. We attribute this to insufficient augmentation during training, as larger performance gains emerge when 2× or 4× augmentation is applied (see Table \ref{tab:area2bump_multi_aug} in main text).

\begin{table}[H]
    \centering
    \begin{tabular}{c|c c c}
    \toprule
     Method     &  Co-smoothing bps($\uparrow$)  &  Behavior decoding($\uparrow$)  &  PSTH $R^2$($\uparrow$) \\
     \midrule
     GRU        &  0.179/$\mathbf{0.204}$(+14.0\%) &   0.698/$\mathbf{0.738}$(+5.7\%) &  0.496/$\mathbf{0.560}$(+12.9\%) \\
     SLDS        &  0.173/$\mathbf{0.182}$(+5.2\%) & 0.732/\textbf{0.763}(+4.2\%) & 0.578/$\mathbf{0.588}$(+1.7\%)  \\
     LFADS       &  0.195/$\mathbf{0.205}$(+5.1\%) & 0.879/$\mathbf{0.882}$(+0.3\%) & 0.552/$\mathbf{0.589}$(+6.7\%)  \\
     NDT        &  0.106/$\mathbf{0.151}$(+42.5\%)  &  0.512/$\mathbf{0.559}$(+9.2\%) & 0.321/$\mathbf{0.343}$(+6.9\%)   \\
     \midrule
     AutoLFADS   & 0.225/$\mathbf{0.231}$(+2.7\%) &  0.882/$\mathbf{0.893}$(+1.2\%) & 0.519/$\mathbf{0.613}$(+1.8\%)  \\
     NDT(ray)   &  0.208/$0.204$(-1.9\%)  &  0.794/$\mathbf{0.807}$(+1.6\%) & 0.414/$\mathbf{0.498}$(+20.3)\%        \\
     \bottomrule
    \end{tabular}
    \caption{\textbf{Metrics before and after EAG augmentation on Area2\_Bump}. Metrics follow the NLB evaluation and are formatted as before/after (improvement ratio). Bold indicates improvements after EAG augmentation.}
    \label{tab:area2bump_aug}
\end{table}

\vspace{2em}
\subsection{S.12\hspace{1em}Supplementary tables on ablation study}
\vspace{1em}

The dimension of random noise can also significantly impact the model performance. We evaluated noise dimensions ($\mathbf{d}_{\text{noise}}$) of 16, 32, 64, 128, and 256, and found that the 64-dimensional setting consistently produced the best results, as shown in Table \ref{tab:ablat_noise}. Consequently, we adopt 64-dimensional noise in EAG.

\begin{table}[h]
    \centering
    \begin{tabular}{c|c c c c}
    \toprule
    $\mathbf{d}_{\text{noise}}$ & $D_{KL}$ psch & RMSE pairwise corr & RMSE mean isi & RMSE std isi \\
    \midrule
    16  & 0.0020 $\pm$ 6.9e-5 & 0.0026 $\pm$ 6.0e-6 & 0.027 $\pm$ 0.005  & 0.019 $\pm$ 0.0006 \\
    32 & 0.0017 $\pm$ 2.9e-4 & 0.0025 $\pm$ 2.1e-5 & 0.035 $\pm$ 0.009  & 0.020 $\pm$ 0.0020 \\
    64  & 0.0014 $\pm$ 2.0e-4 & 0.0024 $\pm$ 1.0e-5 & 0.024 $\pm$ 0.001  & 0.018 $\pm$ 0.0024 \\
    128 & 0.0016 $\pm$ 1.4e-4 & 0.0024 $\pm$ 7.0e-6 & 0.026 $\pm$ 0.001  & 0.019 $\pm$ 0.0011 \\
    256  &  0.0024 $\pm$ 2.7e-4 & 0.0028 $\pm$ 1.1e-5 & 0.025 $\pm$ 0.002  & 0.023 $\pm$ 0.0006 \\
    \bottomrule
    \end{tabular}
    \caption{\textbf{Ablation on noise dimension.} Effect of varying noise dimension on model performance. Based on the results, we select 64 as the final noise dimension.}
    \label{tab:ablat_noise}
\end{table}

The depth and embedding dimension of the Transformer encoder and decoder have a modest impact on performance, but overall the results remain stable. We evaluated EAG on MC\_Maze under several parameter configurations, with the results summarized in Table \ref{tab:ablat_vit_size}. The generation quality is largely robust to variations in model size. Based on these observations, we selected a depth of 4 and an embedding dimension of 256 as the final architecture for generating the MC\_Maze dataset.


\begin{table}[h]
    \centering
    \begin{tabular}{c c | c c c c}
    \toprule
    Depth & Embed dim  & $D_{KL}$ psch  & RMSE pairwise corr & RMSE mean isi & RMSE std isi \\
    \midrule
    2  &  128  & 0.0020 $\pm$ 2.3e-5 & 0.0026 $\pm$ 3.0e-5 & 0.027 $\pm$ 0.007  & 0.019 $\pm$ 0.0003 \\
    
    4  &  256  & 0.0014 $\pm$ 2.0e-4 & 0.0024 $\pm$ 1.0e-5 & 0.024 $\pm$ 0.001  & 0.018 $\pm$ 0.0024 \\
    
    6  &  384  & 0.0017 $\pm$ 1.0e-4 & 0.0024 $\pm$ 1.0e-5 & 0.026 $\pm$ 0.001  & 0.020 $\pm$ 0.0012 \\
    
    8  &  512  & 0.0022 $\pm$ 2.0e-4 & 0.0025 $\pm$ 1.0e-5 & 0.026 $\pm$ 0.001  & 0.019 $\pm$ 0.0009 \\
    
    12 &  768  &  0.0024 $\pm$ 2.0e-4 & 0.0025 $\pm$ 1.1e-5 & 0.027 $\pm$ 0.002  & 0.021 $\pm$ 0.0006 \\
    \bottomrule
    \end{tabular}
    \caption{\textbf{Ablation on transformer depth and embedding dimensions.} Effect of varying depth and embedding dimensions of ViT encoders and decoders on model performance. Based on the results, we select a depth of 4 and an embedding dimension of 256 as the final architecture.}
    \label{tab:ablat_vit_size}
\end{table}